# Unsupervised learning of multiscale switching dynamical system models from multimodal neural data

DongKyu Kim[1], Han-Lin Hsieh[1] and Maryam M. Shanechi[1,2,3,4,*]

[1] Ming Hsieh Department of Electrical and Computer Engineering, Viterbi School of Engineering, University of Southern California, Los Angeles, CA, United States of America
[2] Neuroscience Graduate Program, University of Southern California, Los Angeles, CA, United States of America
[3] Department of Biomedical Engineering, Viterbi School of Engineering, University of Southern California, Los Angeles, CA, United States of America
[4] Department of Computer Science, Viterbi School of Engineering, University of Southern California, Los Angeles, CA, United States of America
* Corresponding author. E-mail: shanechi@usc.edu

**Abstract**

Neural population activity often exhibits regime-dependent non-stationarity in the form of switching dynamics. Learning accurate switching dynamical system models can reveal how behavior is encoded in neural activity. Existing switching approaches have primarily focused on learning models from a single neural modality, either continuous Gaussian signals such as local field potentials (LFPs) or discrete Poisson signals such as spiking activity. However, multiple neural modalities are often recorded simultaneously to measure different spatiotemporal scales of brain activity, and all these modalities can encode behavior. Moreover, regime labels are typically unavailable in training data, posing a significant challenge for learning models of regime-dependent switching dynamics. These gaps highlight the need for a new unsupervised method that can learn switching dynamical system models for multiscale data and do so without requiring regime labels. We develop a novel unsupervised learning algorithm that learns the parameters of switching multiscale dynamical system models using only multiscale neural observations. Doing so, the algorithm can not only fuse multiscale neural information but also account for regime-dependent switches in multiscale neural dynamics. We demonstrate our method using both simulations and two distinct experimental datasets with multimodal spike-LFP observations during different motor tasks. We find that our switching multiscale dynamical system models more accurately decode behavior than switching single-scale dynamical models, showing the success of multiscale neural fusion. Further, our models outperform stationary multiscale models, illustrating the importance of tracking regime-dependent non-stationarity in multimodal neural data. The developed unsupervised learning framework enables more accurate modeling of complex multiscale neural dynamics by leveraging information in multimodal recordings while incorporating regime switches. This approach holds promise for improving the performance and robustness of brain-computer interfaces over time and for advancing our understanding of the neural basis of behavior.

## 1. Introduction

Behavior is encoded in neural population activity across multiple spatiotemporal scales that are measured with different neural modalities [1–7] such as local field potentials (LFP), electrocorticogram (ECoG), and spiking activity [1–13]. As such, a critical problem is to develop dynamical modeling and decoding methods that can fuse information across multimodal neural signals. Field potential signals, such as LFP and ECoG, are commonly modeled as Gaussian processes [2,4,13–15], while spiking activity is typically modeled as a Poisson point process [9,16,17]. To integrate these different modalities, prior work has introduced multiscale dynamical system models for spike-field observations and corresponding multiscale filters (MSF) that decode behavioral states by fusing these modalities [2,3,18]. However, these methods assume that multiscale neural time-series are stationary over time, whereas neural activity often exhibits regime-dependent, non-stationary dynamics [19–23]. Here we address this challenge by developing unsupervised switching dynamical system models for multiscale neural time-series that not only fuse multiscale information but also account for regime-dependent switches in neural dynamics.

One major challenge to developing such models is that the regime state itself is in most cases latent and unobserved. This necessitates the development of algorithms that can learn these models purely based on neural data and unsupervised with respect to the underlying regime, i.e., without any regime labels in the training data. Indeed, while our recent work [20] has developed a switching multiscale filter (sMSF) and a switching multiscale smoother (SMS) for decoding, it has not addressed the major problem of unsupervised learning. Instead, this prior work assumed that the underlying regime states are known in the training set during model learning, which is not the case in most neural datasets.

Given the lack of availability of regime labels in many datasets, developing unsupervised learning algorithms for switching dynamical systems is essential, but remains elusive for multiscale neural data. Indeed, all prior unsupervised switching dynamical modeling methods were for single-scale neural data rather than multiscale data. For example, prior work on single-scale data has developed unsupervised learning algorithms for switching linear dynamical system (SLDS) models with Gaussian observations using the expectation-maximization (EM) algorithm [24]. Furthermore, to enable unsupervised learning of switching dynamical systems models for single-scale Poisson observations, prior work has combined the EM algorithm with a variational Laplace approximation technique to develop the variational Laplace-EM [23]. More recent work has replaced the Laplace approximation with a cubature approximation, leading to a switching Poisson cubature filter (sPCF) with an EM framework [19]. However, none of these unsupervised learning methods are applicable to multiscale neural time-series such as spike-field timeseries.

Here we address the challenges of multiscale neural fusion and unsupervised switching dynamical system modeling simultaneously. We develop an unsupervised method that can learn a switching dynamical system model for multiscale neural observations, purely using the neural data and without any regime labels in the training set. Furthermore, our method can fuse information across the multiscale spike-field modalities to improve the decoding of neural and behavioral states. To develop our learning algorithm, we derive a novel switching multiscale filter, termed the switching multiscale numerical integration filter (sMSNF), and embed it within an EM framework to enable accurate unsupervised learning of model parameters from spike-field observations. To show the importance of the numerical integration-based filter for learning, we also develop an alternative multiscale learning algorithm that uses a Laplace approximation instead and show that the numerical integration approach improves learning and state decoding for the multiscale time-series.

We validate our approach through numerical simulations and two publicly available multiscale spike-LFP motor cortical datasets recorded from nonhuman-primates (NHPs) performing reaching tasks. In simulations, where we have access to ground-truth regimes, we find that the sMSNF-based learning method yields significantly more accurate behavior decoding, neural self-predictions, and regime decoding compared to stationary multiscale methods as well as single-scale switching methods. Furthermore, in experimental datasets, we demonstrate that incorporating switches and combining information from multiple modalities improve the behavior decoding performance. These results highlight the potential of the new method for learning switching multiscale dynamical system models from multimodal neural datasets toward both investigation of neural dynamics and developing brain-computer interfaces (BCIs) that are more accurate and robust over time (see Discussion).

## 2. Methods

In this section, we introduce the switching multiscale dynamical system model (section 2.1) and derive the switching multiscale numerical integration filter (section 2.2). Then, we show how to embed the switching multiscale filter in unsupervised learning frameworks (section 2.3). Finally, we detail the validation methods including numerical simulations (section 2.4) and experimental data analysis (section 2.5).

### 2.1. Switching multiscale dynamical system model

We first present the switching multiscale dynamical system (SMDS) model [20]. Our goal is to learn this model from multiscale neural data and without any regime labels. The regime state $s_t$ is a discrete random variable which specifies the current regime of the system. $s_t$ can take one of $M$ possible values from 1 to $M$, and has Markovian dynamics as follows:

$$P\left(s_t^{(j)} | s_{t-1}^{(i)}\right) = \mathbf{\Phi}_{j,i} \tag{1}$$

where $\mathbf{\Phi}$ is the transition matrix, and $s_t^{(i)}$ and $s_{t-1}^{(j)}$ denote $s_t = i$, and $s_{t-1} = j$ respectively. The elements of $\mathbf{\Phi}$ denote the probability of transitioning from one regime to another regime. For example, $\mathbf{\Phi}_{j,i}$ is the probability of transitioning from regime $i$ to $j$. We also note that $\boldsymbol{\pi}$ is used to denote the initial distribution of regime at $t = 1$, hence $\pi_j = p\left(s_1^{(j)}\right)$.

We model the latent brain state $\mathbf{x}_t$ as a random walk as follows:

$$\mathbf{x}_t = \mathbf{A}(s_t)\mathbf{x}_{t-1} + \mathbf{w}_t, \mathbf{w}_t \sim N\left(0, \mathbf{Q}(s_t)\right) \tag{2}$$

The dynamics matrix $\mathbf{A}(s_t)$ can take one of $M$ possibilities $\{\mathbf{A}^{(1)}, \cdots, \mathbf{A}^{(M)}\}$ depending on the regime, where we denote $\mathbf{A}^{(i)} = \mathbf{A}(s_t^{(i)})$. The same notation convention holds for all system parameters that can be regime state dependent. $\mathbf{w}_t$ is the zero mean white Gaussian noise with the covariance matrix $\mathbf{Q}(s_t)$.

To model the spiking activities, $\mathbf{n}_t$, we use a Poisson observation model with the latent brain states encoding the instantaneous firing rates as follows:

$$f(\mathbf{n}_t | \mathbf{x}_t, s_t) = \prod_{c=1}^{C} \frac{(\lambda_c(\mathbf{x}_t, s_t)\Delta)^{n_t^c} \exp(-\lambda_c(\mathbf{x}_t, s_t)\Delta)}{n_t^c!}$$

$$\lambda_c(\mathbf{x}_t, s_t)\Delta = \exp(\alpha_c(s_t) + \boldsymbol{\beta}_c(s_t)^T \mathbf{x}_t) \tag{3}$$

Specifically, we assume for neuron $c$, among $C$ neurons, the instantaneous firing rate $\lambda_c(\mathbf{x}_t, s_t)$ encodes the latent brain state $\mathbf{x}_t$ with spike model parameters $\alpha_c(s_t)$, and $\boldsymbol{\beta}_c(s_t)$ [2,3,18,25–27]. Then we assume the number of spikes that the neuron $c$ fires $n_t^c$ within a time bin $\Delta$ is Poisson distributed with the rate $\lambda_c(\mathbf{x}_t, s_t)$. Similar to prior work and neural data analyses [3,16,26,28], we assume spiking activities of neurons are conditionally independent of each other conditioned on the brain state and the regime state, and we denote the collection of spikes of all $C$ neurons at time $t$ as $\mathbf{n}_t = [n_t^1, \ldots, n_t^C]^T$.

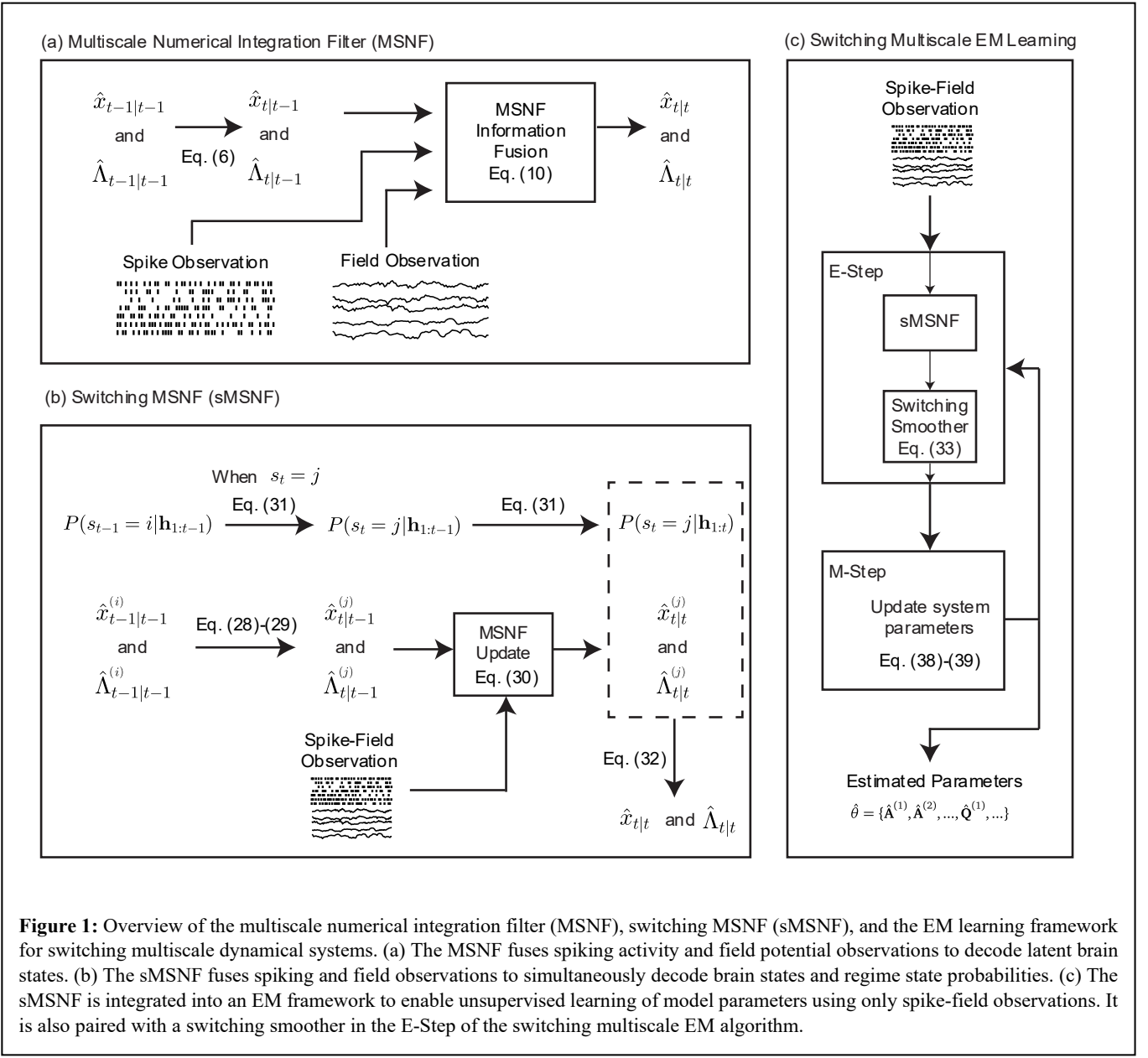

**Figure 1:** Overview of the multiscale numerical integration filter (MSNF), switching MSNF (sMSNF), and the EM learning framework for switching multiscale dynamical systems. (a) The MSNF fuses spiking activity and field potential observations to decode latent brain states. (b) The sMSNF fuses spiking and field observations to simultaneously decode brain states and regime state probabilities. (c) The sMSNF is integrated into an EM framework to enable unsupervised learning of model parameters using only spike-field observations. It is also paired with a switching smoother in the E-Step of the switching multiscale EM algorithm.

We model the field signals, $\mathbf{y}_t$, such as LFP and ECoG with a linear Gaussian model [2–5,29] with an observation matrix $\mathbf{C}(s_t)$, and a noise $\mathbf{r}_t$ which is the zero mean white Gaussian noise with the covariance matrix $\mathbf{R}(s_t)$.

$$\mathbf{y}_t = \mathbf{C}(s_t)\mathbf{x}_t + \mathbf{r}_t, \mathbf{r}_t \sim N\big(0, \mathbf{R}(s_t)\big) \quad (4)$$

We combine the two observation modalities into a multiscale observation modality using a joint likelihood density. Prior studies [2,3,18,20,30] have shown that we can model the field features and spiking activities to be conditionally independent given the latent brain state and the regime state. So, we write the joint likelihood density as:

$$f(\mathbf{h}_t|\mathbf{x}_t, s_t) = f(\mathbf{n}_t, \mathbf{y}_t|\mathbf{x}_t, s_t) = f(\mathbf{n}_t|\mathbf{x}_t, s_t)f(\mathbf{y}_t|\mathbf{x}_t, s_t) \quad (5)$$

where $f(\mathbf{n}_t|\mathbf{x}_t, s_t)$ is the likelihood distribution of spiking activities in (3), and $f(\mathbf{y}_t|\mathbf{x}_t, s_t)$ is the likelihood distribution of field potential signals in (4). We denote the multiscale spike-field observation at time $t$ as $\mathbf{h}_t = [\mathbf{n_t}, \mathbf{y_t}]^T$. Together (1)-(5) form the switching multiscale dynamical system model. We highlight that when $M = 1$, this forms a non-switching multiscale dynamical system model [2,3,18]. To distinguish more clearly, we will refer to $M = 1$ case as the stationary multiscale dynamical system model.

## 2.2. Switching multiscale numerical integration filter (sMSNF)

In this section, we derive the novel switching multiscale numerical integration filter (sMSNF), which incorporates a numerical integration technique to enhance the state inference/decoding. First, we derive a multiscale numerical integration filter (MSNF) that estimates the latent states for stationary multiscale dynamical system models. Then, we extend this filter to incorporate switches (sMSNF).

### 2.2.1. Multiscale Numerical Integration Filter (MSNF)

In this section, we introduce the multiscale numerical integration filter (MSNF) for stationary multiscale dynamical system models as overviewed by Figure 1a. The goal of MSNF is to estimate the unobserved latent state $\mathbf{x}_t$ using multiscale observations $\mathbf{h}_{1:t}$ up to the present time $t$ in a recursive fashion, meaning compute the estimate at time $t$ from that at time $t-1$ recursively. Since there is only 1 regime for the stationary case ($s_t = 1$ for all $t$), we omit the superscript indicating the regime state and denote $\mathbf{A} = \mathbf{A}^{(1)} = \mathbf{A}\big(s_t^{(1)}\big)$.

Generally, a filter tracks the posterior density $f(\mathbf{x}_t|\mathbf{h}_{1:t})$ recursively in time. Since our spiking model in (3) is not Gaussian but Poisson, $f(\mathbf{x}_t|\mathbf{h}_{1:t})$ is not Gaussian. However, as common in various approximations methods [16,23,26], for tractability and computational efficiency, we approximate $f(\mathbf{x}_t|\mathbf{h}_{1:t})$ as Gaussian so we only need to track its mean $\hat{\mathbf{x}}_{t|t}$ and covariance $\widehat{\boldsymbol{\Lambda}}_{t|t}$. Note that the subscript in $\hat{\mathbf{s}}_{i|j}$ for any statistic $\mathbf{s}$ denotes that the statistic is estimated at time $i$ based on neural observations up to time $j$. Now, we develop MSNF to recursively compute $\hat{\mathbf{x}}_{t|t}$ and $\widehat{\boldsymbol{\Lambda}}_{t|t}$ from $\hat{\mathbf{x}}_{t-1|t-1}$ and $\widehat{\boldsymbol{\Lambda}}_{t-1|t-1}$ at time $t-1$. To do so, we first denote the mean and the covariance of the prediction density $f(\mathbf{x}_t|\mathbf{h}_{1:t-1})$ as $\hat{\mathbf{x}}_{t|t-1}$ and $\widehat{\boldsymbol{\Lambda}}_{t|t-1}$, respectively. From (2), the MSNF prediction step is

$$\hat{\mathbf{x}}_{t|t-1} = \mathbf{A}\hat{\mathbf{x}}_{t-1|t-1} \qquad (6)$$
$$\widehat{\boldsymbol{\Lambda}}_{t|t-1} = \mathbf{A}\widehat{\boldsymbol{\Lambda}}_{t-1|t-1}\mathbf{A}^T + \mathbf{Q}$$

Now, we update the latent state estimates by incorporating the new multiscale observation, $\mathbf{h}_t$, using the Bayes rule as follow.

$$\begin{aligned} f(\mathbf{x}_t|\mathbf{h}_{1:t}) &= \frac{f(\mathbf{h}_t|\mathbf{x}_t,\mathbf{h}_{1:t-1})f(\mathbf{x}_t|\mathbf{h}_{1:t-1})}{f(\mathbf{h}_t|\mathbf{h}_{1:t-1})} \\ &= f(\mathbf{n}_t|\mathbf{x}_t,\mathbf{h}_{1:t-1}) \times f(\mathbf{y}_t|\mathbf{x}_t,\mathbf{h}_{1:t-1}) \times \frac{f(\mathbf{x}_t|\mathbf{h}_{1:t-1})}{f(\mathbf{h}_t|\mathbf{h}_{1:t-1})} \\ &= \frac{f(\mathbf{x}_t|\mathbf{n}_t,\mathbf{h}_{1:t-1})f(\mathbf{n}_t|\mathbf{h}_{1:t-1})}{f(\mathbf{x}_t|\mathbf{h}_{1:t-1})} \times \frac{f(\mathbf{x}_t|\mathbf{y}_t,\mathbf{h}_{1:t-1})f(\mathbf{y}_t|\mathbf{h}_{1:t-1})}{f(\mathbf{x}_t|\mathbf{h}_{1:t-1})} \times \frac{f(\mathbf{x}_t|\mathbf{h}_{1:t-1})}{f(\mathbf{h}_t|\mathbf{h}_{1:t-1})} \\ &\propto \frac{f(\mathbf{x}_t|\mathbf{n}_t,\mathbf{h}_{1:t-1})f(\mathbf{x}_t|\mathbf{y}_t,\mathbf{h}_{1:t-1})}{f(\mathbf{x}_t|\mathbf{h}_{1:t-1})} \end{aligned} \qquad (7)$$

The second equality is from the conditional independence between spikes and field signals in (5), and the third equality is from applying the Bayes' rule on each of $f(\mathbf{n}_t|\mathbf{x}_t,\mathbf{h}_{1:t-1})$ and $f(\mathbf{y}_t|\mathbf{x}_t,\mathbf{h}_{1:t-1})$. In the last equation, we only keep the densities that include $\mathbf{x}_t$ because our goal is to track the latent state $\mathbf{x}_t$, and hence use the proportional notation. Again, due to the Poisson observation model (3), $f(\mathbf{x}_t|\mathbf{n}_t,\mathbf{h}_{1:t-1})$ and $f(\mathbf{x}_t|\mathbf{y}_t,\mathbf{h}_{1:t-1})$ are not Gaussian. To compute the latent estimates, similar to many prior works that use a Gaussian approximation on the posterior densities [16,23,26], we approximate $f(\mathbf{x}_t|\mathbf{n}_t,\mathbf{h}_{1:t-1})$ and $f(\mathbf{x}_t|\mathbf{y}_t, \mathbf{h}_{1:t-1})$ as Gaussian with mean and covariance given by $\{\hat{\mathbf{x}}_{\mathbf{n}_{t|t}}, \hat{\mathbf{\Lambda}}_{\mathbf{n}_{t|t}}\}$ and $\{\hat{\mathbf{x}}_{\mathbf{y}_{t|t}}, \hat{\mathbf{\Lambda}}_{\mathbf{y}_{t|t}}\}$, respectively. Then under these Gaussian densities, we expand (7) by focusing on the $\mathbf{x}_t$-related parts (i.e., the exponential function in the Gaussian distribution).

$$f(\mathbf{x}_t|\mathbf{h}_{1:t}) \propto \frac{f(\mathbf{x}_t|\mathbf{n}_t, \mathbf{h}_{1:t-1})f(\mathbf{x}_t|\mathbf{y}_t, \mathbf{h}_{1:t-1})}{f(\mathbf{x}_t|\mathbf{h}_{1:t-1})} \quad (8)$$

$$\propto \exp\left(-\frac{1}{2}\left(\left(\mathbf{x}_t - \hat{\mathbf{x}}_{\mathbf{n}_{t|t}}\right)^T \hat{\mathbf{\Lambda}}_{\mathbf{n}_{t|t}}^{-1} \left(\mathbf{x}_t - \hat{\mathbf{x}}_{\mathbf{n}_{t|t}}\right) + \left(\mathbf{x}_t - \hat{\mathbf{x}}_{\mathbf{y}_{t|t}}\right)^T \hat{\mathbf{\Lambda}}_{\mathbf{y}_{t|t}}^{-1} \left(\mathbf{x}_t - \hat{\mathbf{x}}_{\mathbf{y}_{t|t}}\right)\right.\right.$$

$$\left.\left. - \left(\mathbf{x}_t - \hat{\mathbf{x}}_{t|t-1}\right)^T \hat{\mathbf{\Lambda}}_{t|t-1}^{-1} \left(\mathbf{x}_t - \hat{\mathbf{x}}_{t|t-1}\right)\right)\right)$$

$$\propto \exp\left(-\frac{1}{2}\mathbf{x}_t^T \left(\hat{\mathbf{\Lambda}}_{\mathbf{n}_{t|t}}^{-1} + \hat{\mathbf{\Lambda}}_{\mathbf{y}_{t|t}}^{-1} - \hat{\mathbf{\Lambda}}_{t|t-1}^{-1}\right)\mathbf{x}_t + \mathbf{x}_t^T\left(\hat{\mathbf{\Lambda}}_{\mathbf{n}_{t|t}}^{-1}\hat{\mathbf{x}}_{\mathbf{n}_{t|t}} + \hat{\mathbf{\Lambda}}_{\mathbf{y}_{t|t}}^{-1}\hat{\mathbf{x}}_{\mathbf{y}_{t|t}} - \hat{\mathbf{\Lambda}}_{t|t-1}^{-1}\hat{\mathbf{x}}_{t|t-1}\right)\right)$$

Note that we only keep the $\mathbf{x}_t$-related terms in (8) and so use the proportional relation. Since the Gaussian belongs to the exponential family, the canonical form of $f(\mathbf{x}_t|\mathbf{h}_{1:t})$ is $\exp\left(-\frac{1}{2}\mathbf{x}_t^T(\hat{\mathbf{\Lambda}}_{t|t}^{-1})\mathbf{x}_t + \mathbf{x}_t^T\hat{\mathbf{\Lambda}}_{t|t}^{-1}\hat{\mathbf{x}}_{t|t}\right) \times constant$ [31]. Therefore, we have

$$\hat{\mathbf{\Lambda}}_{t|t}^{-1} = \hat{\mathbf{\Lambda}}_{\mathbf{n}_{t|t}}^{-1} + \hat{\mathbf{\Lambda}}_{\mathbf{y}_{t|t}}^{-1} - \hat{\mathbf{\Lambda}}_{t|t-1}^{-1} \quad (9)$$

$$\hat{\mathbf{\Lambda}}_{t|t}^{-1}\hat{\mathbf{x}}_{t|t} = \hat{\mathbf{\Lambda}}_{\mathbf{n}_{t|t}}^{-1}\hat{\mathbf{x}}_{\mathbf{n}_{t|t}} + \hat{\mathbf{\Lambda}}_{\mathbf{y}_{t|t}}^{-1}\hat{\mathbf{x}}_{\mathbf{y}_{t|t}} - \hat{\mathbf{\Lambda}}_{t|t-1}^{-1}\hat{\mathbf{x}}_{t|t-1}$$

The final step is computing the $\{\hat{\mathbf{x}}_{\mathbf{n}_{t|t}}, \hat{\mathbf{\Lambda}}_{\mathbf{n}_{t|t}}\}$ and $\{\hat{\mathbf{x}}_{\mathbf{y}_{t|t}}, \hat{\mathbf{\Lambda}}_{\mathbf{y}_{t|t}}\}$ from prediction $\{\hat{\mathbf{x}}_{t|t-1}, \hat{\mathbf{\Lambda}}_{t|t-1}\}$. To do so, we can use the update equations in the Poisson cubature filter (PCF) [19] and the Kalman filter (KF) [32], respectively. We provide the update equations of these two filters in appendix 7.1. The final update equation of MSNF is as follows (details are in appendix 7.2).

$$\hat{\mathbf{\Lambda}}_{t|t}^{-1} = \hat{\mathbf{\Lambda}}_{t|t-1}^{-1} + \tilde{\mathbf{C}}^T\tilde{\mathbf{R}}^{-1}\tilde{\mathbf{C}} + \mathbf{C}^T\mathbf{R}^{-1}\mathbf{C} \quad (10)$$

$$\hat{\mathbf{x}}_{t|t} = \hat{\mathbf{x}}_{t|t-1} + \hat{\mathbf{\Lambda}}_{t|t}\tilde{\mathbf{C}}^T\tilde{\mathbf{R}}^{-1}[\mathbf{n}_t - \hat{\mathbf{n}}_{t|t-1}] + \hat{\mathbf{\Lambda}}_{t|t}\mathbf{C}^T\mathbf{R}^{-1}[\mathbf{y}_t - \mathbf{C}\hat{\mathbf{x}}_{t|t-1}]$$

Because field signals are sampled at slower rates than spikes, we model the field signals as missing observations when spiking activities are observed, but new field signals are not available. We achieve this by setting the observation matrix $\mathbf{C} = 0$ at those time-steps. Together, (6) and (10) define the MSNF for stationary multiscale dynamical systems.

### 2.2.2. Extension of MSNF to switching multiscale dynamical systems

Having developed MSNF for stationary systems, we now extend it to handle switching dynamics as overviewed in Figure 1b. Here, the goal is to estimate the posterior probability of the regime state $P(s_t|\mathbf{h}_{1:t})$ and the posterior density of the latent brain state $f(\mathbf{x}_t|\mathbf{h}_{1:t})$, which can be rewritten as

$$f(\mathbf{x}_t|\mathbf{h}_{1:t}) = \sum_{j=1}^{M} f\left(\mathbf{x}_t|\mathbf{h}_{1:t}, s_t^{(j)}\right) \times P\left(s_t^{(j)}|\mathbf{h}_{1:t}\right) \quad (11)$$

Again, we approximate $f\left(\mathbf{x}_t|\mathbf{h}_{1:t}, s_t^{(j)}\right)$ as Gaussian whose mean and covariance are $\hat{\mathbf{x}}_{t|t}^{(j)}$ and $\hat{\mathbf{\Lambda}}_{t|t}^{(j)}$, respectively. Therefore, the goal is to compute $\hat{\mathbf{x}}_{t|t}^{(j)}, \hat{\mathbf{\Lambda}}_{t|t}^{(j)}$, and $P\left(s_t^{(j)}|\mathbf{h}_{1:t}\right)$ from time $t-1$ to $t$ recursively. In our prior work with supervised learning, we developed a switching multiscale filter (sMSF) [20] that used the Laplace approximation to compute the recursions. However, here we are interested in unsupervised learning, for which the Laplace approximation is not sufficiently accurate as we will show. As such, to enable accurate unsupervised learning, here the key insight is that we can replace the Laplace approximation with the more accurate numerical integration approach from our new MSNF. Computing the regime state posteriors $P\left(s_t^{(j)}\middle|\mathbf{h}_{1:t}\right)$ from $P\left(s_{t-1}^{(i)}\middle|\mathbf{h}_{1:t-1}\right)$ follows the same process as in sMSF [20]. However, we need to change the recursive flow of computing the brain state posterior $f\left(\mathbf{x}_t\middle|\mathbf{h}_{1:t}, s_t^{(j)}\right)$ to include our new numerical integration method. To see this, note that in sMSF, the brain state posterior computation follows the below step-wise flow:

$$f\left(\mathbf{x}_{t-1}\middle|\mathbf{h}_{1:t-1}, s_{t-1}^{(i)}\right) \xrightarrow{\text{step 1}} f\left(\mathbf{x}_{t-1}\middle|\mathbf{h}_{1:t-1}, s_t^{(j)}\right) \xrightarrow{\text{step 2}} f\left(\mathbf{x}_t\middle|\mathbf{h}_{1:t-1}, s_t^{(j)}\right) \xrightarrow{\text{step 3}} f\left(\mathbf{x}_t\middle|\mathbf{h}_{1:t}, s_t^{(j)}\right) \quad (12)$$

In the above, steps 2 and 3 are the prediction and update steps of the latent brain state $\mathbf{x}_t$, respectively. Therefore, to compute these with numerical integration instead of Laplace, we can use (6) and (10) in MSNF for these two steps, while just conditioning everything on the regime state $s_t^{(j)}$. We term this extension of MSNF for switching multiscale dynamical system models as sMSNF. For completeness, we provide equations for the sMSNF in appendix 7.3.

In our previous work [20], we also derived the switching multiscale smoother (SMS) that computes non-causal estimations of the latent brain states and the regime states. The smoother only requires the outputs of the switching filters regardless of the estimation method (whether Laplace based or numerical integration based), and thus can be used as presented in [20] here as well. We provide the equations of SMS in appendix 7.4.

### 2.2.3. Likelihood scaling for model mismatch between different modalities

When combining continuous and discrete neural signal modalities, a fundamental challenge arises. Spikes follow Poisson statistics while field potentials follow Gaussian statistics. These different probabilistic characteristics can create imbalanced contributions during filtering, where one modality may dominate the estimation process regardless of the information content it carries [33].

To address this scaling mismatch, we introduce a likelihood weighting parameter $\tau$ that adjusts the relative contribution of the Gaussian field observations and Poisson point process spike observations. Since MSNF is a special case of sMSNF with $M = 1$, we adopt the sMSNF notation for all related equations. We can write the multiscale likelihood function as

$$\log f(\mathbf{h}_t|\mathbf{x}_t, s_t) = \log f(\mathbf{n}_t|\mathbf{x}_t, s_t) + \tau \log f(\mathbf{y}_t|\mathbf{x}_t, s_t) \quad (13)$$

This parameter is determined empirically through grid search in cross-validation to optimize behavioral decoding performance. It is easy to see that the scaling factor modifies the filter equations by weighing the field observation terms as follows:

$$\left(\widehat{\mathbf{\Lambda}}_{t|t}^{(j)}\right)^{-1} = \left(\widehat{\mathbf{\Lambda}}_{t|t-1}^{(j)}\right)^{-1} + \widetilde{\mathbf{C}}^{(j)^T}\widetilde{\mathbf{R}}^{(j)^{-1}}\widetilde{\mathbf{C}}^{(j)} + \tau \mathbf{C}^{(j)^T}\mathbf{R}^{(j)^{-1}}\mathbf{C}^{(j)}$$

$$\widehat{\mathbf{x}}_{t|t}^{(j)} = \widehat{\mathbf{x}}_{t|t-1}^{(j)} + \widehat{\mathbf{\Lambda}}_{t|t}^{(j)}\widetilde{\mathbf{C}}^{(j)^T}\widetilde{\mathbf{R}}^{(j)^{-1}}\left[\mathbf{n}_t - \widehat{\mathbf{n}}_{t|t-1}^{(j)}\right] + \tau \widehat{\mathbf{\Lambda}}_{t|t}^{(j)}\mathbf{C}^{(j)^T}\mathbf{R}^{(j)^{-1}}\left[\mathbf{y}_t - \mathbf{C}^{(j)}\widehat{\mathbf{x}}_{t|t-1}^{(j)}\right] \quad (14)$$

where the $\tau$ scaling specifically affects the field potential contribution while leaving the spike contribution unmodified. The details of the scaling parameter's propagation are presented in appendix 7.5.

### 2.3. Unsupervised learning of switching multiscale dynamical system model with EM

To learn the parameters of switching multiscale dynamical systems without regime labels, we can embed our switching filters (either sMSF or sMSNF) in an expectation-maximization (EM) framework, as illustrated in Figure 1c. This unsupervised switching multiscale EM is distinct from our previous framework that instead performed supervised learning and assumed that regime labels are known in the training data [20]. The set of parameters to be learned is as follows:

$$\theta = \left\{ \left\{ \mathbf{A}^{(j)}, \mathbf{Q}^{(j)}, \left\{ \alpha_c^{(j)}, \boldsymbol{\beta}_c^{(j)} \right\}_{c=1:C}, \mathbf{C}^{(j)}, \mathbf{R}^{(j)}, \pi_j, \left\{ \boldsymbol{\Phi}_{j,i} \right\}_{i=1:M} \right\}_{j=1:M}, \boldsymbol{\mu}_0, \boldsymbol{\Lambda}_0 \right\}, \quad (15)$$

The learned parameter $\hat{\theta} = \underset{\theta}{\operatorname{argmax}} \log f(\mathbf{h}_{1:T}; \theta)$ should maximizes the observation log-likelihood given $\mathbf{h}_{1:T}$. However, computing $\log f(\mathbf{h}_{1:T}; \theta)$ directly is not tractable from the dynamical model in general, so EM finds a lower bound of it as follows:

$$\log f(\mathbf{h}_{1:T}; \theta) = \log \left( \sum_{s_T=1}^{M} \cdots \sum_{s_1=1}^{M} \int \left[ \frac{q(\mathbf{x}_{0:T}, s_{1:T}; \theta) \times f(\mathbf{x}_{0:T}, s_{1:T}, \mathbf{h}_{1:T}; \theta)}{q(\mathbf{x}_{0:T}, s_{1:T}; \theta)} \right] d\mathbf{x}_{0:T} \right)$$
$$\geq \sum_{s_T=1}^{M} \cdots \sum_{s_1=1}^{M} \int q(\mathbf{x}_{0:T}, s_{1:T}; \theta) \times \log \left[ \frac{f(\mathbf{x}_{0:T}, s_{1:T}, \mathbf{h}_{1:T}; \theta)}{q(\mathbf{x}_{0:T}, s_{1:T}; \theta)} \right] d\mathbf{x}_{0:T} \quad (16)$$

This evidence lower bound (ELBO) comes from the Jensen's inequality [34]. Therefore, EM is an iterative method that alternates between an expectation step (E-step) and a maximization step (M-step) to increase the ELBO in each iteration until convergence [35,36]. If we are in iteration $k-1$, the E-step involves using the current set of estimated system parameters $\theta^{(k-1)}$ to find the optimal $q(\mathbf{x}_{0:T}, s_{1:T}; \theta^{(k-1)}) = f(\mathbf{x}_{0:T}, s_{1:T} | \mathbf{h}_{1:T}; \theta^{(k-1)})$, which is the smooth density of the latent state and the regime state. We compute this density using our new numerical integration filter sMSNF and its smoothing version mentioned above. After the E-step, we focus on the M-step, which maximizes ELBO given $q(\mathbf{x}_{0:T}, s_{1:T}; \theta^{(k-1)})$ by finding the optimal parameter set $\theta^{(k)}$. $\theta^{(k)}$ becomes the new set of estimated system parameters and EM proceeds to the next iteration.

We now more explicitly write the EM equations. We first write the complete data log-likelihood function for the latent brain states, the regime states, and the multiscale observations as follows:

$$\log f(\mathbf{x}_{0:T}, s_{1:T}, \mathbf{h}_{1:T}; \theta) = \log P(s_{1:T}; \theta) + \log f(\mathbf{x}_{0:T} | s_{1:T}; \theta) + \log f(\mathbf{h}_{1:T} | \mathbf{x}_{0:T}, s_{1:T}; \theta)$$
$$= \log P(s_1) + \sum_{t=2}^{T} \log P(s_t | s_{t-1}) + \log N(\mathbf{x}_0; \boldsymbol{\mu}_0, \boldsymbol{\Lambda}_0) + \sum_{t=1}^{T} \log f(\mathbf{x}_t | \mathbf{x}_{t-1}, s_t)$$
$$+ \sum_{t=1}^{T} \sum_{c=1}^{C} \log P(n_t^c | \mathbf{x}_t, s_t) + \tau \sum_{t=1}^{T} \log f(\mathbf{y}_t | \mathbf{x}_t, s_t)$$
$$P\left(s_1^{(j)}\right) = \pi_j$$
$$P\left(s_t^{(j)} \middle| s_{t-1}^{(i)}\right) = \boldsymbol{\Phi}_{j,i} \quad (17)$$
$$f\left(\mathbf{x}_t \middle| \mathbf{x}_{t-1}, s_t^{(j)}\right) = N\left(\mathbf{x}_t; \mathbf{A}^{(j)} \mathbf{x}_{t-1}, \mathbf{Q}^{(j)}\right)$$
$$P\left(n_t^c \middle| \mathbf{x}_t, s_t^{(j)}\right) = Poisson\left(n_t^c; \exp\left(\alpha_c^{(j)} + \boldsymbol{\beta}_c^{(j)^T} \mathbf{x}_t\right)\right)$$
$$f\left(\mathbf{y}_t \middle| \mathbf{x}_t, s_t^{(j)}\right) = N\left(\mathbf{y}_t; \mathbf{C}^{(j)} \mathbf{x}_t, \mathbf{R}^{(j)}\right)$$

For brevity, all the terms after the first equality are given the same parameter set $\theta$. $N(\mathbf{x}; \boldsymbol{\mu}, \boldsymbol{\Lambda})$ denotes a multivariate normal probability density function with mean $\boldsymbol{\mu}$, and covariance $\boldsymbol{\Lambda}$, and $Poisson(n; \lambda)$ denotes the Poisson probability mass function with parameter $\lambda$. $\tau$ is the likelihood scaling parameter discussed in section 2.2.3.

The expected complete data log-likelihood function in the ELBO (16) for the latent brain states, the regime states, and spike-field observations in our switching multiscale dynamical model is as follows [18,19,24]:

$$
\begin{aligned}
&E\left[\log f(\mathbf{x}_{0:T}, s_{1:T}, \mathbf{h}_{1:T}; \theta^*) | \mathbf{h}_{1:T}; \theta^{(k-1)}\right] \\
&= \sum_{j=1}^{M} P\left(s_1^{(j)} | \mathbf{h}_{1:T}\right) \log \pi_j + \sum_{t=2}^{T} \sum_{i,j=1}^{M} P\left(s_t^{(j)}, s_{t-1}^{(i)} | \mathbf{h}_{1:T}\right) \log \boldsymbol{\Phi}_{j,i} \\
&+ E\left[\log N(\mathbf{x}_0; \boldsymbol{\mu}_0, \boldsymbol{\Lambda}_0) | \mathbf{h}_{1:T}; \theta^{(k-1)}\right] \\
&+ \sum_{t=1}^{T} \sum_{j=1}^{M} P\left(s_t^{(j)} | \mathbf{h}_{1:T}\right) E\left[\log N\left(\mathbf{x}_t; \mathbf{A}^{(j)} \mathbf{x}_{t-1}, \mathbf{Q}^{(j)}\right) | \mathbf{h}_{1:T}, s_t^{(j)}; \theta^{(k-1)}\right] \\
&+ \sum_{t=1}^{T} \sum_{j=1}^{M} P\left(s_t^{(j)} | \mathbf{h}_{1:T}\right) \\
&\times \sum_{c=1}^{C} E\left[\log Poisson\left(\mathrm{n}_t^c; \exp\left(\alpha_c^{(j)} + \boldsymbol{\beta}_c^{(j)^T} \mathbf{x}_t\right)\right) | \mathbf{h}_{1:T}, s_t^{(j)}; \theta^{(k-1)}\right] \\
&+ \tau \sum_{t=1}^{T} \sum_{j=1}^{M} P\left(s_t^{(j)} | \mathbf{h}_{1:T}\right) E\left[\log N\left(\mathbf{y}_t; \mathbf{C}^{(j)} \mathbf{x}_t, \mathbf{R}^{(j)}\right) | \mathbf{h}_{1:T}, s_t^{(j)}; \theta^{(k-1)}\right]
\end{aligned}
\quad (18)
$$

The E-step can compute the above expected log-likelihood using our new numerical integration filter sMSNF and smoother. The details of using smoothed states to compute the expected log-likelihood is in appendix 7.6. Then, the M-step finds the $\theta^{(k)}$ that maximizes this expected log-likelihood:

$$
\theta^{(k)} = \arg\max_{\theta^*} E\left[\log f(\mathbf{x}_{0:T}, s_{1:T}, \mathbf{h}_{1:T}; \theta^*) | \mathbf{h}_{1:T}; \theta^{(k-1)}\right] \quad (19)
$$

We note that, similar to prior work [19,20], since the parameters associated with each regime are independent of those in other regimes, the system effectively decouples into separate models when conditioned on a single regime. Consequently, the expected complete-data log-likelihood can be decomposed by regime, making the M-step of the EM algorithm separable. This allows optimization to be performed independently for each regime. For example, the dynamics matrix for regime $j$, denoted $\mathbf{A}^{(j)}$, that maximizes (19) is:

$$
A^{(j)} = \left(\sum_{t=1}^{T} P\left(s_t^{(j)} | \mathbf{h}_{1:T}\right) E\left[\mathbf{x}_t \mathbf{x}_{t-1}^T | \mathbf{h}_{1:T}, s_t^{(j)}\right]\right) \\
\times \left(\sum_{t=1}^{T} P\left(s_t^{(j)} | \mathbf{h}_{1:T}\right) E\left[\mathbf{x}_{t-1} \mathbf{x}_{t-1}^T | \mathbf{h}_{1:T}, s_t^{(j)}\right]\right)^{-1} \quad (20)
$$

All equations for the M-step are provided in appendix 7.7. We refer to the switching EM framework using our Laplace-based sMSF as sMSF-EM and to the EM framework using our new numerical integration-based filter sMSNF as sMSNF-EM. The stationary counterparts, in which the number of regimes is fixed to $M = 1$, are denoted MSF-EM and MSNF-EM, respectively. Note that both sMSF-EM and sMSNF-EM are novel and have not been previously introduced or analyzed. By comparing these two unsupervised learning frameworks here, we will show the importance of the numerical integration-based filter for unsupervised multiscale learning.

Table 1 summarizes the various method names for convenience.

### 2.4. Numerical simulation framework

We first validate the proposed unsupervised switching multiscale EM framework through numerical simulations. Specifically, we simulate spiking activity and continuous field features using randomly generated SMDS models. The learning algorithms are evaluated by training models on synthetic training datasets and testing them on independently generated test datasets. Performance is assessed using the evaluation metrics in Section 2.4.2.

To evaluate the performance of our model in the stationary (non-switching) case first, we simulate data using SMDS model with only 1 regime, and learn models using MSF-EM [18] and our MSNF-EM in order to show the benefits of using numerical integration in multiscale data fusion/filtering even for stationary systems. In addition, we compare against single modality models in this stationary case, whether single-modal Gaussian or single-modal Poisson point processes. Specifically, we use the Kalman filter embedded in EM for learning with single-modality Gaussian observations, and use both the point process filter (PPF) [16,18,26,27] and the Poisson Cubature Filter (PCF) [19] embedded in EM for learning with single-modality Poisson point process observations. We term these methods KF-EM, PPF-EM, and PCF-EM, respectively. All evaluations are done in cross-validation with independent test data.

To evaluate the performance of our model in the switching systems, we simulate data using SMDS model with 2 regimes, and we learn models using sMSF-EM and sMSNF-EM, as well as single modality algorithms, switching Kalman filter with EM (sKF-EM) [21,24], switching PPF with EM (sPPF-EM) [20], and switching PCF with EM (sPCF-EM) [19]. Similar to the stationary systems, we test all the learned models using separately simulated testing data. We provide a summary of all the methods in Table 1.

**Table 1** Summary of all the methods used for evaluation. Abbreviation, name, switching capabilities, and observation type that each method can handle are listed.

| Abbreviation | Method Name | Switching | Observation Type |
| --- | --- | --- | --- |
| KF-EM | Kalman Filter-EM | No | Gaussian |
| PPF-EM | Point Process Filter-EM | No | Poisson |
| PCF-EM | Poisson Cubature Filter-EM | No | Poisson |
| MSF-EM | Multiscale Filter-EM | No | Multiscale[a] |
| MSNF-EM (new) | Multiscale Numerical Integration Filter-EM | No | Multiscale[a] |
| sKF-EM | Switching Kalman Filter-EM | Yes | Gaussian |
| sPPF-EM | Switching Point Process Filter-EM | Yes | Poisson |
| sPCF-EM | Switching Poisson Cubature Filter-EM | Yes | Poisson |
| sMSF-EM (new) | Switching Multiscale Filter-EM | Yes | Multiscale[a] |
| sMSNF-EM (new) | Switching Multiscale Numerical Integration Filter-EM | Yes | Multiscale[a] |

[a]Multiscale indicates both Gaussian and Poisson.

### 2.4.1. Numerical simulation settings

For numerical simulations, we randomly form switching multiscale dynamical systems (SMDS) and simulate data from them. Motivated by experimental datasets, across all simulations, we assume the time bin to be $\Delta = 10$ms. For each simulation setting, we form 30 separate systems with a training set consisting of 10k timesteps, and a separate testing set consisting also of 10k timesteps. For stationary simulations, we set the number of regimes to be 1, and for switching simulations, we set the number of regimes to be 2 and set the probability of staying within the same regime to $\Phi_{jj} = 0.99$, and the probability of switching to the other regime to $\Phi_{ij} = 0.01$ for all pairs of $j \neq i$.

For the latent brain state dynamic equation parameters, we set the dimension of the latent brain states $x_t$ to 10, $\dim(\mathbf{x}_t) \triangleq d$. We further randomly form the dynamics matrix and noise covariance matrix. Prior work [7] suggests that there exist both shared and distinct dynamical modes in spiking activity and field features. Using the procedures described in [7], we assume the eigenvalues of **A** are stable and comprised of $d/2$ complex conjugate pairs $\{r_i e^{\pm j\theta_i}\}_{i \in [1,d/2]}$. $r_i$ is randomly chosen between [0.99,0.995] and $\theta_i$ is chosen between [0,0.063] uniformly. For our simulation, we allocate 3 pairs of eigenvalues to be shared, and 1 pair of unshared eigenvalues to each of the modalities. We also randomly select the eigenvalues of **Q** to be between [0.01, 0.04]. For switching systems, we follow the same procedure for all regimes.

For Poisson observations, we let the number of neurons to be 30 and let each neuron to have a base firing rate between [6, 9] Hz and a maximum firing rate between [40, 50] Hz. Then we randomly select parameters based on the base and the maximum firing rates and generate Poisson observations using (3).

For field observations, we set the dimension of the field observations to be 30, the range of values to be between [26,30], signal-to-noise ratio (SNR) to be between [0.3,0.35], and set $\mathbf{y}_t$ to be available every 50ms. The signal-to-noise ratio for $f$th feature is defined to be $SNR_f = std\left(\left[\frac{\mathbf{c}_f^{(s_t)} \mathbf{x}_t}{\sqrt{R_f^{(s_t)}}}\right]_{t \in T}\right)$, where $std(\cdot)$ denotes the standard deviation, and $\mathbf{C}_f^{(s_t)}$ denotes the $f$th row of $\mathbf{C}^{(s_t)}$, whose values depend on the regime state $s_t$ at time $t$. In addition, $R_f^{(s_t)}$ is the noise variance for $f$th feature given regime $s_t$. For our simulations, we assume the field observation noise variance matrices $\mathbf{R}^{(j)}$ are diagonal. Then we randomly select parameters and generate field observations using (4). Similar to the latent dynamics, for switching systems, we follow the same procedure for all regimes to have a separate set of observation parameters per regime.

### 2.4.2. Learning and performance metrics

After simulating both stationary and switching datasets, we learn system models using all stationary methods (KF-EM, PPF-EM, PCF-EM, MSF-EM, MSNF-EM; Table 1) on the stationary dataset and using all switching methods (sKF-EM, sPPF-EM, sPCF-EM, sMSF-EM, and sMSNF-EM; Table 1) on the switching dataset to validate and show the advantage of our new unsupervised learning method (MSNF-EM and sMSNF-EM) in both stationary and switching cases.

We initialize the parameters randomly for each system, except for the latent state dynamic matrix $A$, which we set to $0.9 * \mathbf{I}$. We set the initial transition matrices such that the probability of staying in the same regime is $\mathbf{\Phi}_{jj} = 0.995$. For fair comparisons among stationary methods, and among switching models, we initialize the models with the same set of parameters. Once parameters are learned for each model, we evaluate the models on a test dataset that has been generated separately.

To evaluate the performance of the learned models in simulations, we use the latent state decoding correlation coefficient (CC) metric. We calculate the CC between the estimated latent brain states and the true latent brain states for each dimension and then average over dimensions. However, since the latent states are latent, and there exists an infinite number of equivalent latent state space models within a similarity transform [29], we need to perform a similarity transform before directly comparing the latent brain states. To do so, we follow the procedure described in [29] to find the similarity transform that minimizes the mean-squared error between the predicted latent brain states and true latent brain states, and then compare them (see details in [29]).For both stationary and switching systems, we also find the latent state decoding CC using the true model parameters, $CC_{true}$, and learnt model parameters with each learning algorithm (Table 1). We then compute a normalized CC metric given by $CC/CC_{true}$ for each learning method, showing the learnt model's comparison to the ideal case where model parameters are exactly known.

For switching systems, we also evaluate the estimated regime state $\hat{s}_t = \hat{s}_{t|t} = \underset{j \in [1,M]}{\operatorname{argmax}} P(s_t^{(j)}|\mathbf{h}_{1:t})$ by computing the regime accuracy:

$$Accuracy(s_{1:T}, \hat{s}_{1:T}) = \sum_{t=1}^{T} [s_t = \hat{s}_t]/T \quad (21)$$

Since our learning method is unsupervised, the assigned number for the learned regime may not match the true regime label, so we calculate the accuracy over all permutations of regimes $1:M$, and choose the maximum accuracy for each method.

We also evaluate the neural self-predictions of field features. For switching systems, we compute this metric by obtaining the one-step-ahead prediction of the latent brain states for each regime, and combining them using the regime probability as follows:

$$\hat{\mathbf{y}}_{t|t-1} = \sum_{j=1}^{M} \mathbf{C}^{(j)} \hat{\mathbf{x}}_{t|t-1}^{(j)} P(s_t^{(j)}|\mathbf{h}_{1:t-1}) \quad (22)$$

Then we calculate the CC between the predicted field features and the true field features and average the CCs over dimensions of the field observations. We also obtain the CC using the true system parameters and use it to normalize the CC computed from learned parameters as $CC/CC_{true}$. For stationary systems, we compute the field prediction by $\hat{\mathbf{y}}_{t|t-1} = \mathbf{C}\hat{\mathbf{x}}_{t|t-1}$, and apply the same procedure to obtain the prediction CC, and the normalized prediction CC.

The last metric that we obtain is the neural self-prediction metric of Poisson observations. We use the procedure outlined in [19] to compute the accuracy of one-step-ahead prediction of spiking activity through the probability $P(n_t^c \geq 1|\mathbf{h}_{1:t-1})$. We use $P(n_t^c \geq 1|\mathbf{h}_{1:t-1})$ per neuron to classify if a spike was observed for the time bin $t$, and obtain an area under the curve (AUC) measure. Then we average the AUC measures over all neurons and normalize the average AUC to yield predictive power by $PP = 2 * AUC - 1$ such that 0 is chance and 1 is perfect prediction [7,19]. We also compute this value using the true system parameters and use it to normalize the $PP$ of the learned system as $PP/PP_{true}$.

## 2.5. Experimental data

We further validate the learning algorithms using two publicly available nonhuman-primate (NHP) datasets. The overview of the behavior task in the two datasets are illustrated in Figure 2. The first dataset contains recordings from the primary motor cortex (M1) of an NHP (NHP I) performing 2D continuous reaches for targets in a square grid using a cursor in a virtual reality environment from the Sabes Lab [37,38]. The animal was presented with an 8-by-8 square grid in a virtual reality environment and was trained to make 2D reaches for targets using a controllable cursor by using its fingertip on its left arm. The target is considered acquired if the animal holds on to the target within an acceptance zone for 450ms. A new target shows up in a random position on the grid upon the acquisition of the previous target. Further details can be found in [37,38]. All animal procedures were performed in accordance with the U. S. National Research Council's Guide for the Care and Use of Laboratory Animals and were approved by the UCSF Institutional Animal Care and Use Committee. We refer to the first dataset as the grid reaching NHP dataset.

The second dataset contains recordings from the primary motor cortex (M1) of an NHP (NHP C), performing an eight-target, center-out target reach task on a screen using a cursor controlled by a two link manipulandum. The animal was tasked to reach one of the eight target squares that are distributed around a circle, after holding on at the center of the circle for 0.5-0.6s. Target reach is considered successful if the animal reaches the target within 1.5s and holds for a random time between 0.2s and 0.4s.

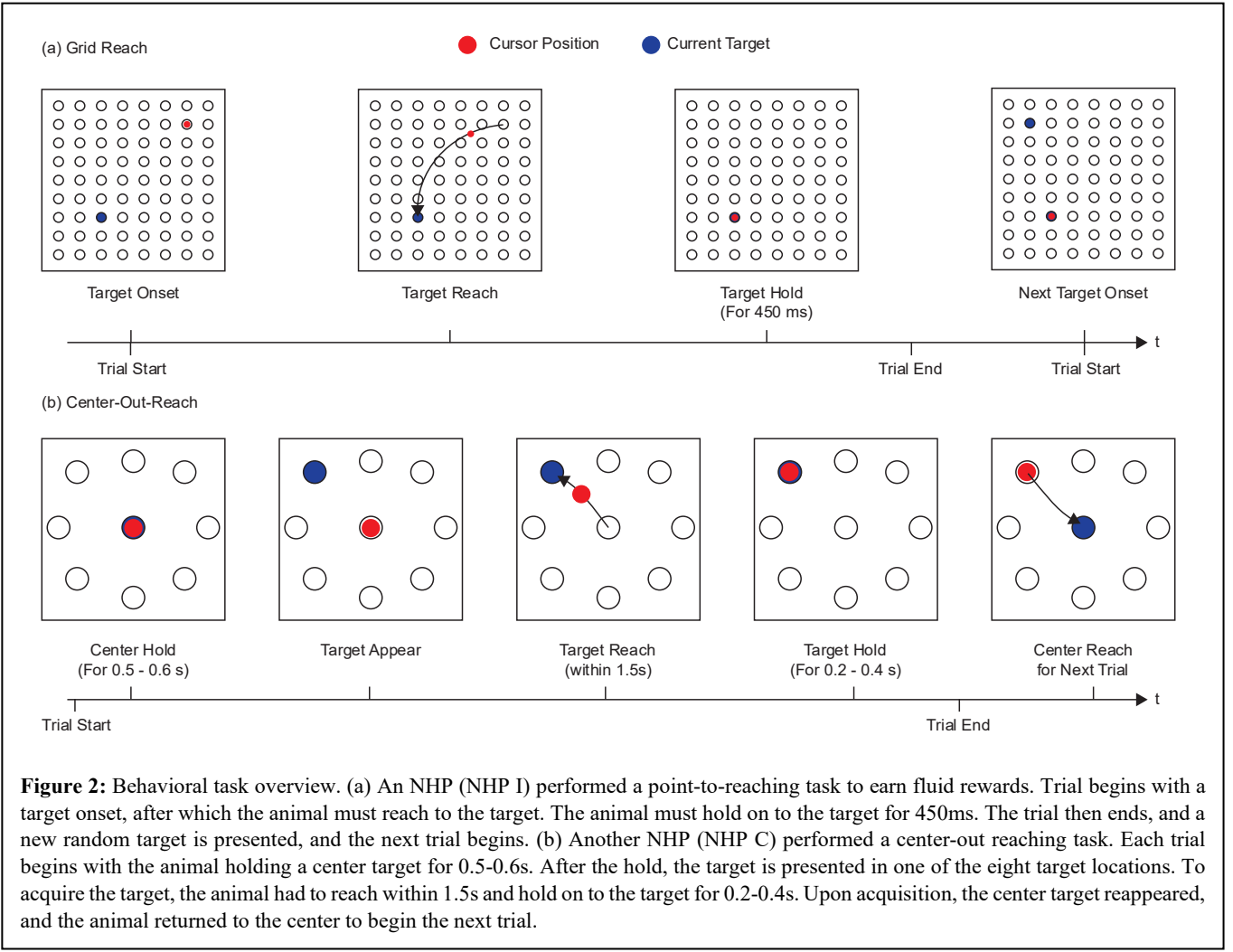

**Figure 2:** Behavioral task overview. (a) An NHP (NHP I) performed a point-to-reaching task to earn fluid rewards. Trial begins with a target onset, after which the animal must reach to the target. The animal must hold on to the target for 450ms. The trial then ends, and a new random target is presented, and the next trial begins. (b) Another NHP (NHP C) performed a center-out reaching task. Each trial begins with the animal holding a center target for 0.5-0.6s. After the hold, the target is presented in one of the eight target locations. To acquire the target, the animal had to reach within 1.5s and hold on to the target for 0.2-0.4s. Upon acquisition, the center target reappeared, and the animal returned to the center to begin the next trial.

Further details of the experiment can be found in [14]. All animal procedures were performed with approval from the Institutional Animal Care and Use Committee of Northwestern University. We refer to the second dataset as the center-out-reach NHP dataset.

*2.5.1. Data preprocessing*

Neural activity for both datasets were recorded from a 96-channel silicon microelectrode array (Blackrock Microsystems, Salt Lake City, UT). We use the multiunit spike counts from each channel as the discrete neural signals, and we set the bin size to be $\Delta = 10\ ms$. For the grid reaching NHP dataset, we extract LFP power features following the approach used in our previous work [2,3]. After common average referencing, we compute power features in seven standard frequency bands (theta: [4, 8] Hz, alpha: [8, 12] Hz, low beta: [12, 24] Hz, mid-beta: [24, 34] Hz, high beta: [34, 55] Hz, low gamma: [65, 95] Hz, and high gamma: [130, 170] Hz) with short-time Fourier transform with 300ms causal moving windows every 50ms. We analyze data from five sessions (sessions 20160915_01-20160927_06) that had consistent data lengths and both spikes and LFPs available. For the center-out-reach NHP dataset, we extract LFP power features using the method described in [14], in the following frequency bands ([0, 4], [7, 20], [70, 115], [130, 200] and [200, 300] Hz) and use 256ms causal moving windows every 50ms. We analyze data from four sessions (e1_sess1, e2_sess2, e3_sess1, e3_sess2). For both datasets, we analyze each session individually with a five-fold cross-validation.

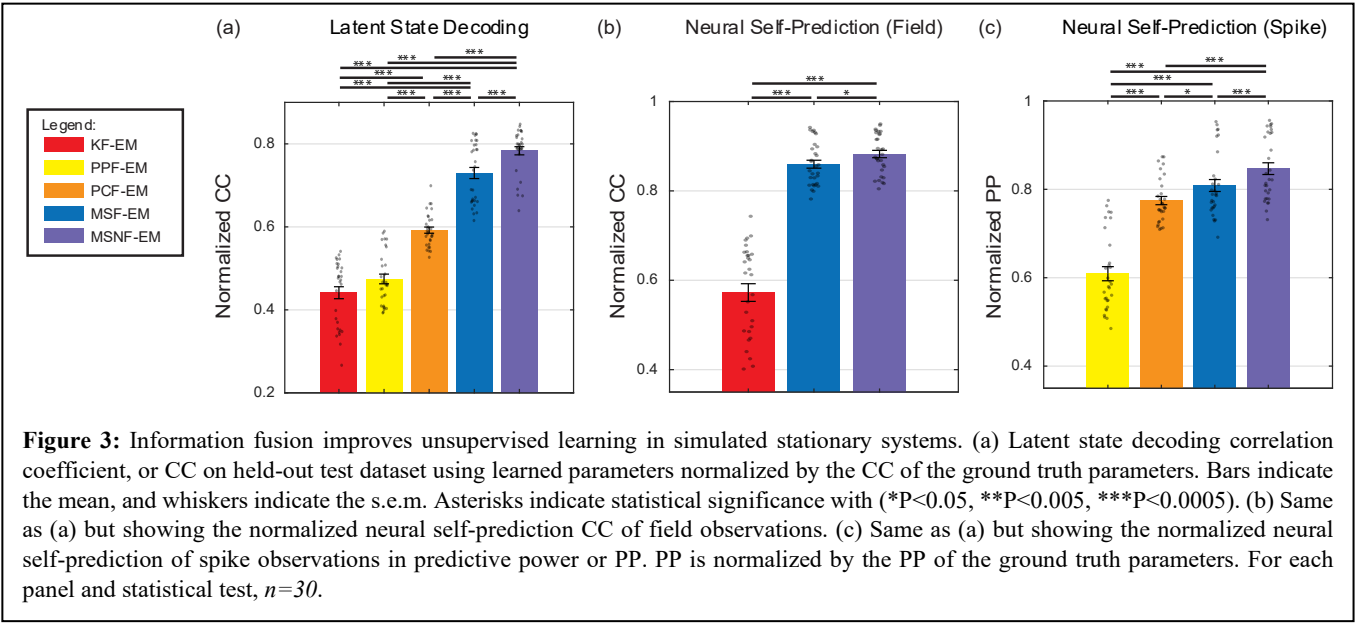

**Figure 3:** Information fusion improves unsupervised learning in simulated stationary systems. (a) Latent state decoding correlation coefficient, or CC on held-out test dataset using learned parameters normalized by the CC of the ground truth parameters. Bars indicate the mean, and whiskers indicate the s.e.m. Asterisks indicate statistical significance with (*P<0.05, **P<0.005, ***P<0.0005). (b) Same as (a) but showing the normalized neural self-prediction CC of field observations. (c) Same as (a) but showing the normalized neural self-prediction of spike observations in predictive power or PP. PP is normalized by the PP of the ground truth parameters. For each panel and statistical test, *n=30*.

*2.5.2. Learning procedure and evaluation*

To first investigate how our new numerical-integration based unsupervised learning framework, sMSNF-EM, fuses information from multiple modalities, we perform a controlled study by gradually including signals of one type while keeping a fixed number of channels of the other type in the model. For example, to examine the effect of fusing spike channels to LFPs, we fix the number of LFP channels (base channels) at 5 and incrementally include 5, 10, and 20 spike channels. We then train models to evaluate how the addition of spike channels influences model performance. We repeat this process several times by randomly selecting different spike and LFP channels in the evaluations. We also repeat this process by changing the number of base LFP channels to 10 and 20. Finally, we study the reverse case in which the base channels are spike channels and by fusing LFP channels to a fixed number of base spike channels (5, 10, 20). This process is again repeated by using many random permutations of the channels in our evaluation.

We first show the benefit of multiscale fusion in our method by training our sMSNF-EM models on all training sets and comparing their performance to single-scale learning methods: sKF-EM for LFPs and sPCF-EM for spikes. We further show the benefit of our multiscale learning by training and comparing the multiscale models (MSF-EM, MSNF-EM, sMSF-EM, sMSNF-EM) on the full training set consisting of all 20 LFP channels and 20 spike channels.

We learn system parameters by running all EM-based methods for 300 iterations to give sufficient time for convergence. We take the dimension of the brain state as $d = 10$, based on common values used in prior works [3,19,29]. For switching methods, based on prior work [19], we set the number of regimes as $M = 2$. As in prior work [19], to concentrate on the changes in the latent brain state dynamics, we learn the same set of neural observation parameters across all regimes.

All the methods used are unsupervised and only the spiking activities and the LFP power features are used to estimate the system parameter. To determine the scaling factor for likelihood balancing between spike-LFP modalities, we use an inner four-fold cross-validation scheme within the training set to learn models of varying scaling factors and pick the scaling factor with the highest behavior decoding CC in this inner cross-validation within the training set. To enable fair comparison across methods, we learn the scaling factor using MSF-EM and apply the resulting value to all multiscale methods.

As our main metric in real data, since we do not know the true latent states, we compute the behavior decoding CC in cross-validation. This is defined as the CC between the decoded behavior by any model and the true behavior (averaged over the dimensions of behavior). For the grid reaching NHP dataset, we regard the behavior as the 2D cursor velocities. For the center-out-reach NHP dataset, we regard the behavior as the 2D cursor positions and velocities. For each model, we linearly project the decoded training set latent states onto the desired behavior in the training set to learn the behavior map from the latent states; we then apply this learnt map on the decoded latent states in the test set for that model. We also evaluate the models using neural self-predictions of field features and Poisson observations, as described in Section 2.4.2.

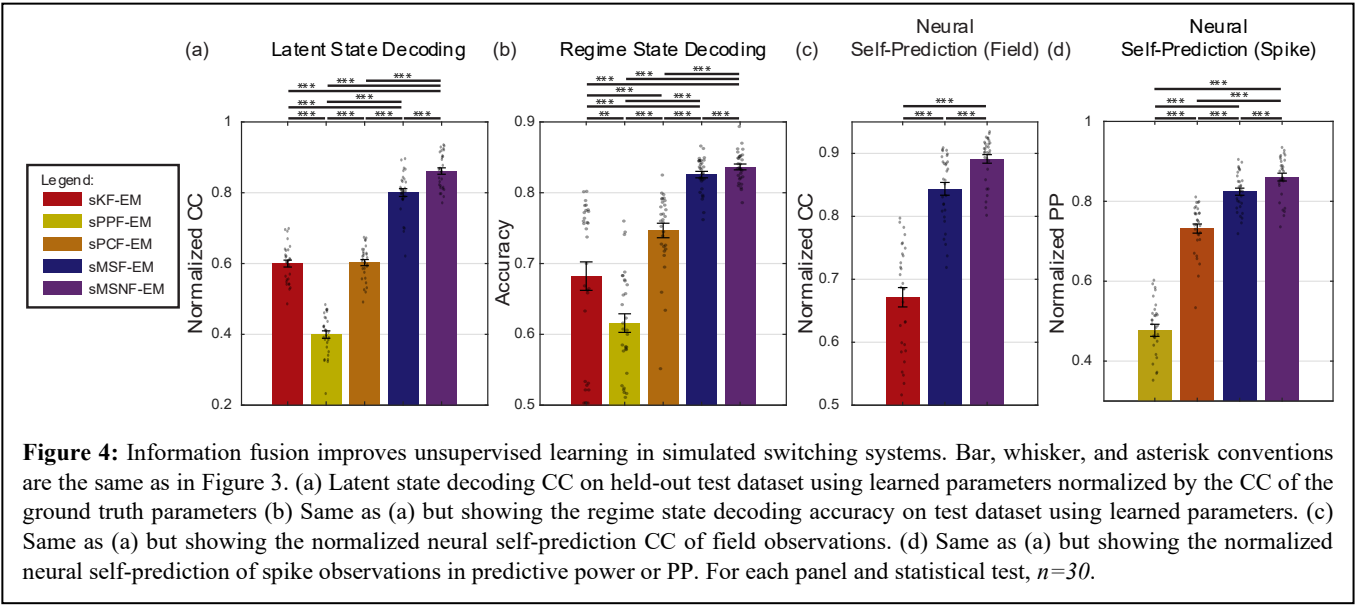

**Figure 4:** Information fusion improves unsupervised learning in simulated switching systems. Bar, whisker, and asterisk conventions are the same as in Figure 3. (a) Latent state decoding CC on held-out test dataset using learned parameters normalized by the CC of the ground truth parameters (b) Same as (a) but showing the regime state decoding accuracy on test dataset using learned parameters. (c) Same as (a) but showing the normalized neural self-prediction CC of field observations. (d) Same as (a) but showing the normalized neural self-prediction of spike observations in predictive power or PP. For each panel and statistical test, *n=30*.

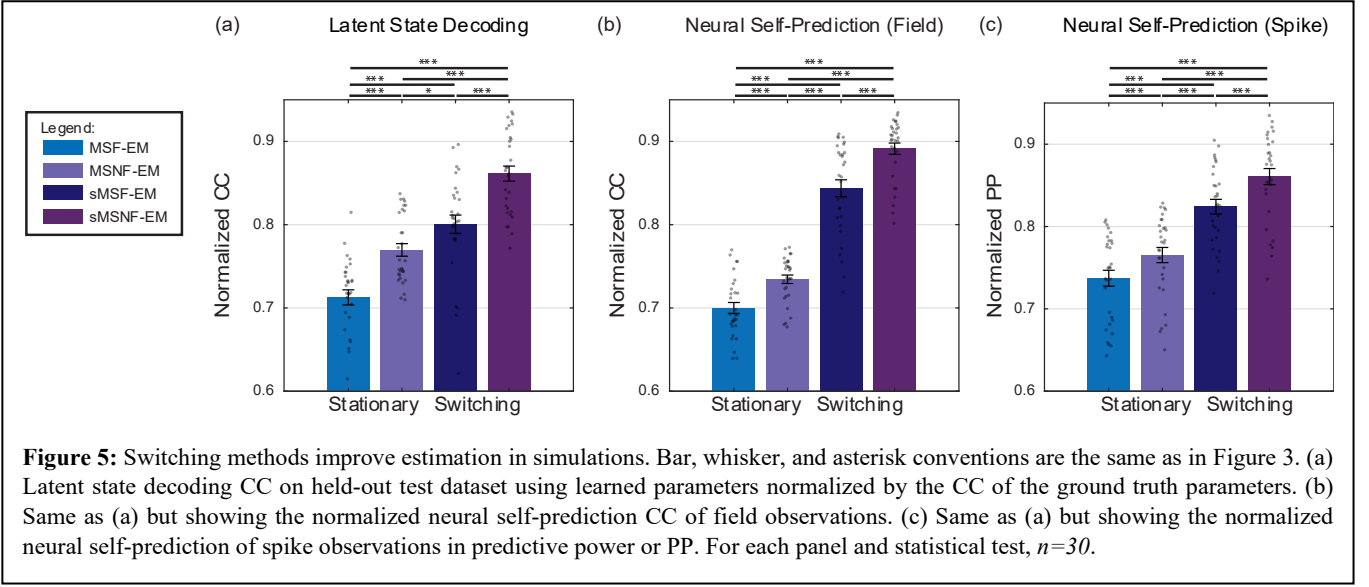

**Figure 5:** Switching methods improve estimation in simulations. Bar, whisker, and asterisk conventions are the same as in Figure 3. (a) Latent state decoding CC on held-out test dataset using learned parameters normalized by the CC of the ground truth parameters. (b) Same as (a) but showing the normalized neural self-prediction CC of field observations. (c) Same as (a) but showing the normalized neural self-prediction of spike observations in predictive power or PP. For each panel and statistical test, *n=30*.

## 3. Results

We show in both numerical simulations and real neural activities recorded from NHPs that the developed unsupervised switching multiscale learning framework is capable of estimating both latent brain states and regime states while fusing information from both Gaussian and Poisson observations. For all statistical comparisons we use the Wilcoxon signed-rank test, and to control for false discovery rate, we use the Benjamini-Hochberg Procedure [39].

*3.1. By enabling multiscale information fusion, our method yields higher accuracy in stationary simulations*

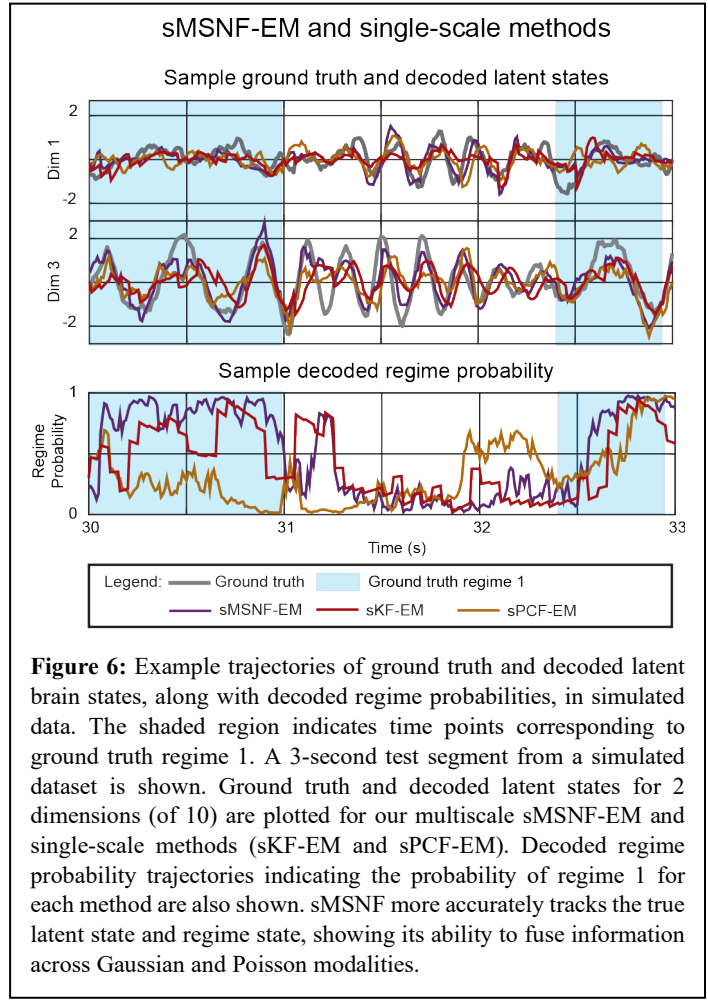

**Figure 6:** Example trajectories of ground truth and decoded latent brain states, along with decoded regime probabilities, in simulated data. The shaded region indicates time points corresponding to ground truth regime 1. A 3-second test segment from a simulated dataset is shown. Ground truth and decoded latent states for 2 dimensions (of 10) are plotted for our multiscale sMSNF-EM and single-scale methods (sKF-EM and sPCF-EM). Decoded regime probability trajectories indicating the probability of regime 1 for each method are also shown. sMSNF more accurately tracks the true latent state and regime state, showing its ability to fuse information across Gaussian and Poisson modalities.

First, we validate that our MSNF-EM can accurately learn system parameters in stationary simulations and effectively fuse information from Gaussian and Poisson observations to improve estimation performance. We find that the latent brain state decoding obtained using our stationary multiscale method significantly outperform those from single-scale methods ($p \leq 2.80\text{e-}6$, $N=30$). As shown in Figure 3a, integrating field features into spike data improves normalized latent state decoding by 32.30% when comparing MSNF-EM to PCF-EM. This shows that information fusion yields better latent state decoding. Similarly, adding spike features to field features enhances decoding performance by 77.53% using MSNF-EM.

We also compared MSNF-EM for stationary systems to MSF-EM for stationary systems that uses the Laplace approximation instead of the numerical integration. We find that the new MSNF-EM outperforms the MSF-EM in latent state decoding ($p \leq 2.80\text{e-}6$, $N=30$), showing the importance of our new method even in the stationary case.

Finally, improvements extend beyond latent state decoding. As shown in Figure 3b and 3c, our multiscale method also yields higher performance in neural self-predictions compared with single-scale models of either spike or field modalities, confirming the advantage of multiscale information fusion ($p \leq 1.38\text{e-}4$, $N=30$). Our MSNF-EM also again outperforms MSF-EM for neural prediction ($p \leq 2.29\text{e-}2$, $N=30$).

### 3.2. By enabling multiscale information fusion, our method yields higher accuracy in switching simulations

To enable unsupervised learning of switching multiscale dynamical systems, here we developed the Laplace-based sMSF-EM, and the numerical integration based sMSNF-EM. We find that multiscale data fusion using either method significantly improves parameter estimation for switching systems in simulations and that sMSNF-EM outperforms sMSF-EM. First, consistent with the results in Section 3.1, both our multiscale frameworks, sMSF-EM and sMSNF-EM, achieve higher decoding performance across evaluation metrics compared to single-scale methods, sKF-EM, sPPF-EM, and sPCF-EM as shown in Figure 4 ($p \leq 2.80\text{e-}6$, $N=30$). Second, our numerical integration-based sMSNF-EM outperforms our Laplace-based sMSF-EM throughout Figure 4 ($p \leq 1.23\text{e-}5$, $N=30$), showing the importance of numerical integration for accurate multiscale unsupervised learning.

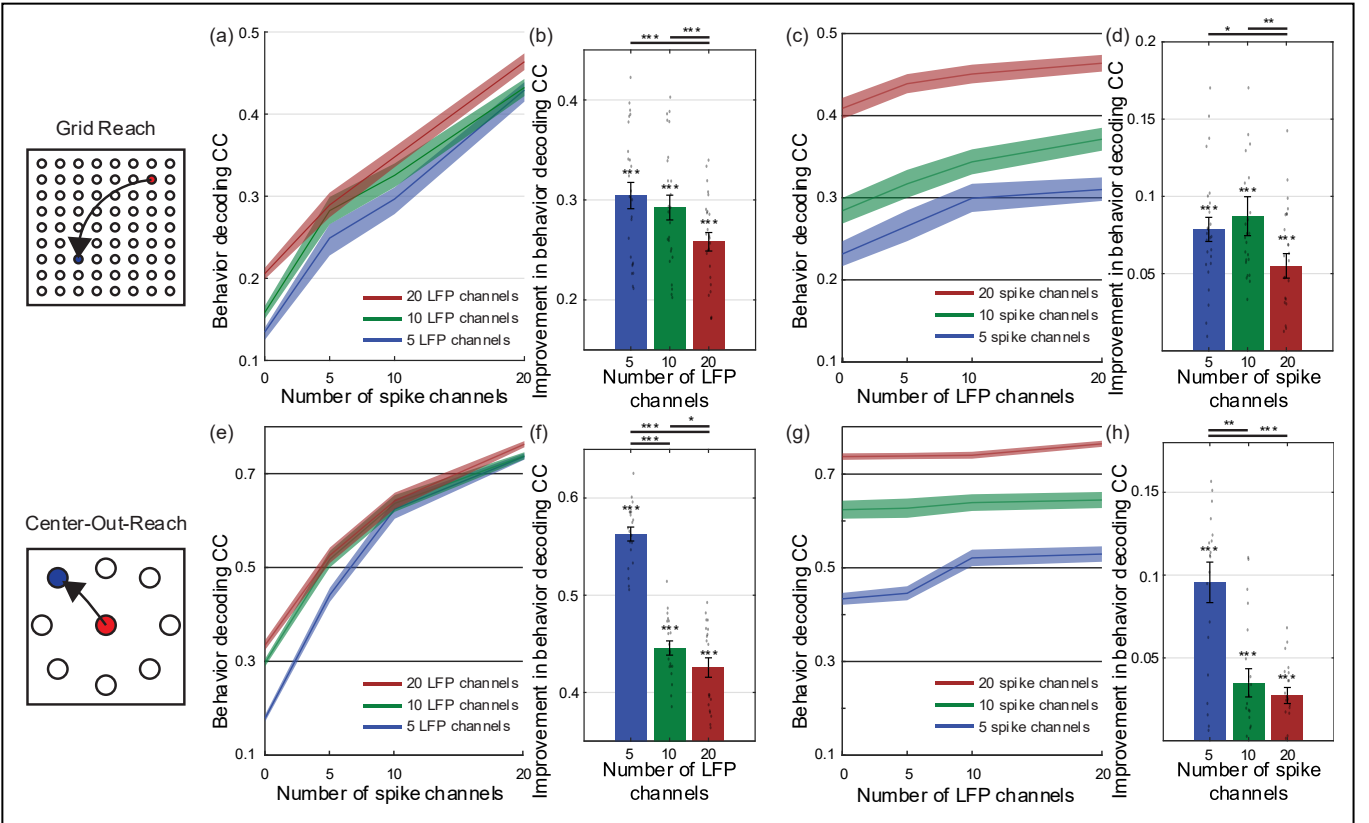

**Figure 7:** In NHP datasets, switching multiscale EM outperforms switching single-scale EM in behavior decoding. Asterisk conventions are the same as in Figure 3. (a) Behavior decoding CC on held-out test dataset using learned parameters as more spike channels are added to the baseline LFP channels for the grid reaching NHP dataset. The start of the curves indicates the behavior decoding CC for the single-scale method using just the baseline channels. The solid line indicates the mean, and the shaded area represents s.e.m. (b) Comparison of the maximum improvement of behavior decoding after combining 20 spike channels with varying numbers of baseline LFP channels. Bar, and whisker conventions are the same as in Figure 3. The asterisks above each bar indicate statistical significance against 0. (c) Same as (a) but now with LFP channels being added to baseline spike channels. (d) Same as (b) but showing improvements for different numbers of baseline spike channels. (e)-(h) Same as (a)-(f) but for the center-out-reach NHP dataset. For each panel and statistical test, $n=25$ for the grid reaching dataset, and $n=20$ for the center-out-reach dataset.

To further verify that capturing regime-dependent non-stationarity helps, we also trained the stationary MSF-EM and MSNF-EM models on the same switching dataset. As shown in Figure 5, incorporating switching dynamics, as enabled by sMSF-EM and sMSNF-EM, leads to significant improvements across all evaluation metrics ($p \leq 7.28\text{e-}3$, $N=30$).

Finally, we visualized the behavior of the switching models by plotting example trajectories of the ground truth and decoded latent brain states and of regime state probabilities in Figure 6. To highlight the benefits of information fusion, we compared the performance of the multiscale method (sMSNF-EM) with single-scale methods (sKF-EM, sPCF-EM). Figure 6 shows that sMSNF-EM more accurately tracks the underlying latent brain states and produces regime probability estimates that better align with the ground truth in the simulation, showing its success in multiscale information fusion.

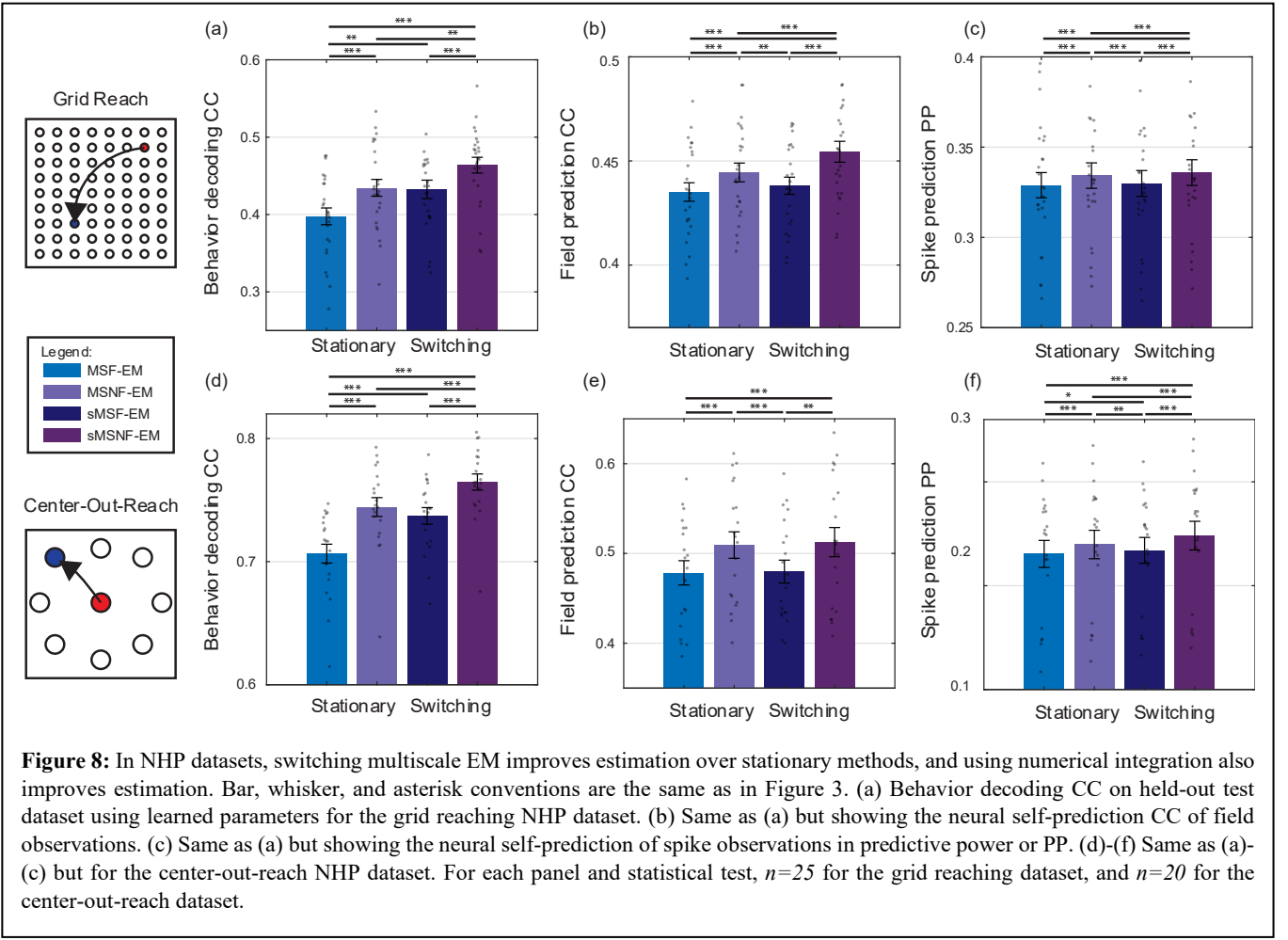

**Figure 8:** In NHP datasets, switching multiscale EM improves estimation over stationary methods, and using numerical integration also improves estimation. Bar, whisker, and asterisk conventions are the same as in Figure 3. (a) Behavior decoding CC on held-out test dataset using learned parameters for the grid reaching NHP dataset. (b) Same as (a) but showing the neural self-prediction CC of field observations. (c) Same as (a) but showing the neural self-prediction of spike observations in predictive power or PP. (d)-(f) Same as (a)-(c) but for the center-out-reach NHP dataset. For each panel and statistical test, *n=25* for the grid reaching dataset, and *n=20* for the center-out-reach dataset.

Collectively, the above improvements demonstrate the success of our method in fusing spike and field observations in switching state-space models and the advantage of doing so for better latent state decoding and neural prediction.

### 3.3. Our method enables better movement decoding in NHP motor cortical spike-LFP data by enabling multiscale information fusion

We evaluate our switching multiscale method, sMSNF-EM, on two publicly available datasets that include both LFP power features and spiking activity. For comparison, we also apply single-scale methods, sKF-EM for LFPs and sPCF-EM for spikes, as baselines. In the first dataset, neural recordings were collected from the NHP motor cortical areas during performance of a 2D reach task toward targets arranged in a square grid within a virtual environment. In the second dataset, recordings were collected from a different animal performing a center-out-reach task using a two-link manipulandum.

We find that behavior decoding performance improves consistently as additional channels from a secondary signal modality are fused with the baseline modality using our method. As shown in Figure 7a,c,e,g, the behavior decoding CCs increase with the addition of LFP channels to spike channels, and vice versa, across both experimental datasets. Moreover, Figure 7b,d,f,h demonstrate that the improvements in behavior performance achieved by multiscale methods are statistically significantly ($p \leq 3.41e-5$, *N=25* for Figure 7b,d, and $p \leq 2.07e-4$, *N=20* for Figure 7f,h). These results provide evidence that our method enables multiscale information fusion and that doing so leads to enhanced behavioral decoding accuracy.

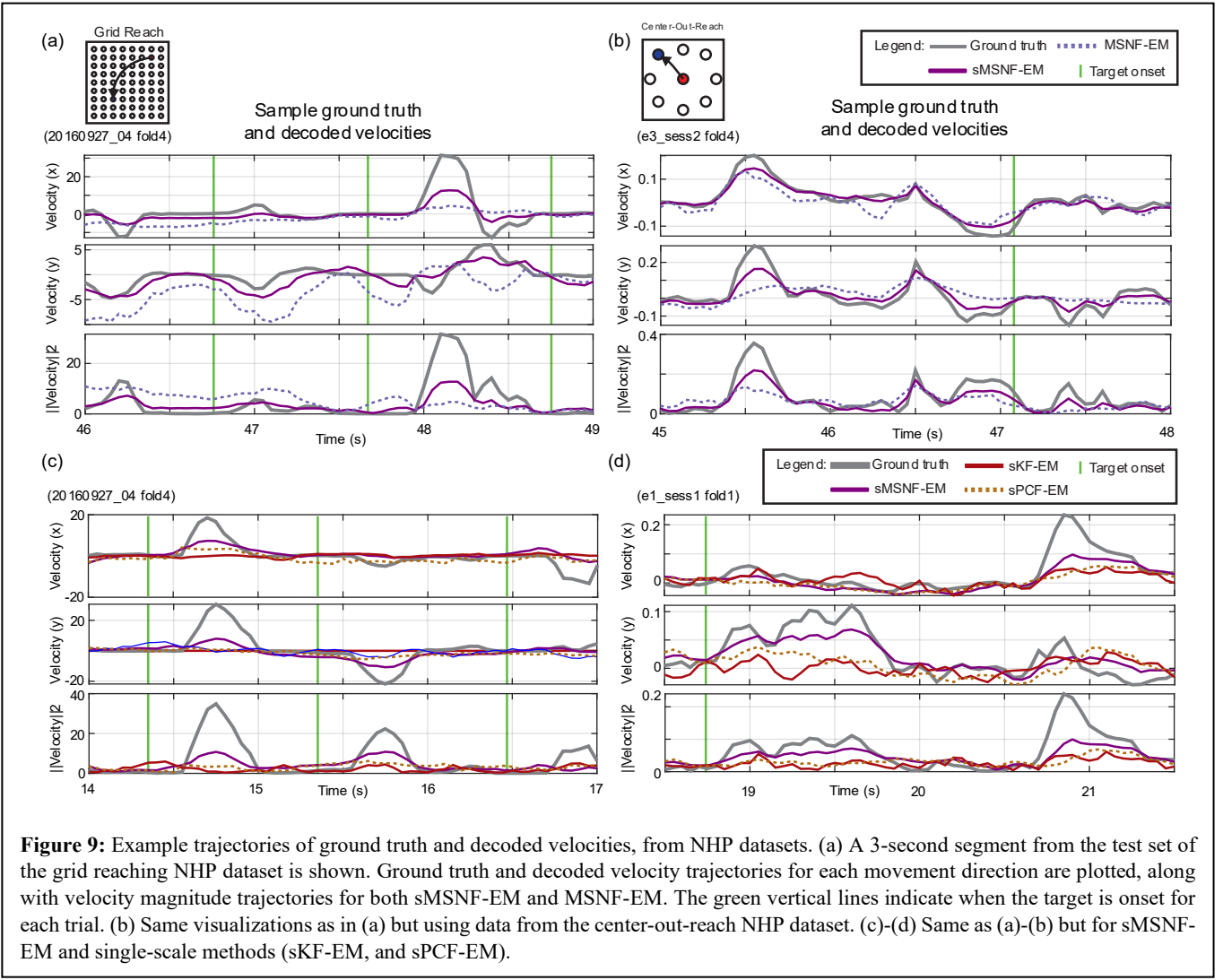

**Figure 9:** Example trajectories of ground truth and decoded velocities, from NHP datasets. (a) A 3-second segment from the test set of the grid reaching NHP dataset is shown. Ground truth and decoded velocity trajectories for each movement direction are plotted, along with velocity magnitude trajectories for both sMSNF-EM and MSNF-EM. The green vertical lines indicate when the target is onset for each trial. (b) Same visualizations as in (a) but using data from the center-out-reach NHP dataset. (c)-(d) Same as (a)-(b) but for sMSNF-EM and single-scale methods (sKF-EM, and sPCF-EM).

### 3.4. Our method achieves better movement decoding in NHP motor cortical spike-LFP data by allowing for switching dynamics

Next, we evaluate all multiscale methods – the stationary MSF-EM and MSNF-EM, and the switching sMSF-EM, and sMSNF-EM (Table 1) – on the same motor cortical datasets. These comparisons enable a systematic investigation of the effects of switching dynamics and the benefits of numerical integration.

As shown in Figure 8a and 8d switching methods achieve significantly higher behavioral decoding CCs than stationary methods. This improvement is consistent across both datasets ($p \leq 7.68e{-}4$, $N=25$ for Figure 8a, and $p \leq 3.10e{-}4$, $N=20$ for Figure 8d). Moreover, numerical integration methods, switching sMSNF-EM and stationary MSNF-EM, consistently outperform their Laplace-based counterparts, MSF-EM and sMSNF-EM, in behavioral decoding. This shows that numerical integration methods lead to more accurate unsupervised learning of multiscale models, regardless of whether switching is enabled. These differences are statistically significant, ($p \leq 2.07e{-}4$, $N=25$ for Figure 8a, and $p \leq 1.40e{-}4$, $N=20$ for Figure 8d), highlighting the benefit of using numerical integration.

Figure 8b,c,e,f further show that numerical integration also leads to improvements in neural self-prediction performance for both LFP power features and spikes ($p \leq 1.57e{-}5$, $N=25$ for Figure 8b,c and $p \leq 1.62e{-}3$, $N=20$ for Figure 8e,f). However, the impact of switching on neural self-prediction is not uniform across tasks. sMSNF-EM yields improved self-prediction of LFP power features compared to MSNF-EM in grid reaching task but not in the center-out reach task.

Finally, Figure 9 provides example visualizations of behavior decoding on both experimental datasets. Figure 9a shows sample trajectories of the ground truth and decoded velocities across movement directions, along with the overall velocity magnitude, for sMSNF-EM and MSNF-EM on the grid-reaching NHP dataset. Figure 9b shows similar visualizations for the center-out-reach NHP dataset. In both cases, we observe that our sMSNF-EM, which uses switching dynamics, more accurately tracks the ground truth behavior compared to our MSNF-EM, which uses stationary dynamics. This highlights the advantage of using switching dynamical systems. In addition, Figure 9c,d show similar visualizations for sMSNF-EM and single-scale methods (sKF-EM, and sPCF-EM). In all these cases, we observe that our multiscale switching dynamical system model tracks the ground turth behavior more accurately, and thus highlights the advantage of multiscale fusion.

## 4. Discussion

In this work, we developed an unsupervised learning framework for switching multiscale dynamical systems, enabling the identification of system parameters from simultaneous Gaussian and Poisson observations without requiring explicit regime labels in the training set. To enhance model accuracy, we introduced a novel switching multiscale filter (sMSNF) that incorporates a numerical integration technique, along with its stationary counterpart (MSNF). We demonstrated how these filters can be embedded within an EM framework to enable unsupervised learning of both stationary and switching dynamical systems with multiscale observations. We showed that our framework can enable better behavior decoding by fusing multiscale spike-field information and by allowing for switching dynamics. Additionally, we developed an alternative unsupervised multiscale learning algorithm using a Laplace-based switching multiscale filter (sMSF), showing that our new numerical integration-based sMSNF filter enables more accurate unsupervised learning for multiscale data.

A robust multiscale system should effectively leverage information from all available modalities, thus achieving multiscale fusion. We conducted analyses showing that our method can fuse multiple modalities while also tracking regime switches. Doing so, our method improved latent state, regime state, and behavioral decoding in simulations. Similarly, in experimental datasets, we observed that our method improves behavior decoding by fusing LFP and spike modalities. A successful switching system should also exploit the ability to transition between regimes. We showed that our method could successfully track regime switches in simulations. In experimental data, incorporating switching capabilities in addition to multiscale fusion, as enabled by our method, again enhanced behavior decoding.

One key contribution of this work is the incorporation of numerical integration into multiscale filters to improve the accuracy of unsupervised learning with multiscale data. We showed that compared with the commonly used Laplace approximation, numerical integration enables more precise unsupervised learning for multiscale Gaussian-Poisson data. Across both simulations and experimental datasets, incorporating numerical integration consistently outperformed using Laplace-based approximations for learning.

The ability of sMSNF-EM to learn switching systems from multiple modalities in an unsupervised manner opens up several promising applications. Such switching models are important for investigating how behavior is represented in neural activity, especially during complex tasks that have different regimes [40,41]. These models are also important for building real-world BCIs that can handle complex and non-stationary tasks with context switches [1,42,43]. In such systems, sMSNF-EM can improve decoding performance compared to stationary or single scale methods, as demonstrated in our experimental results. Furthermore, a major challenge in BCIs is that neural spiking activity may degrade over time due to various factors such as electrode tip encapsulation [44–48]. In comparison, LFPs have been shown to exhibit improved stability over time [46,47]. By fusing information across spiking and LFP activity, our new learning method can also enable more robust BCI systems that allow users to maintain accurate and stable performance over longer time periods, thus facilitating the clinical viability of these BCI systems.

## 5. Acknowledgements

Authors acknowledge support of the ONR award N00014-19-1-2128, NSF CRCNS Award IIS 2113271, and NIH awards R01MH123770 and R61MH135407.

# 7. Appendix

## 7.1. Kalman filter and PCF

Here we present the update equations for the Kalman filter and the PCF which we will use to derive MSNF. The update equation of Kalman filter can be written as follows:

$$\begin{aligned}
\hat{\mathbf{x}}_{\mathbf{y}_{t|t}} &= \hat{\mathbf{x}}_{t|t-1} + \widehat{\mathbf{\Lambda}}_{\mathbf{y}_{t|t}} \mathbf{C}^T \mathbf{R}^{-1} [\mathbf{y}_t - \mathbf{C}\hat{\mathbf{x}}_{t|t-1}] \\
\widehat{\mathbf{\Lambda}}_{\mathbf{y}_{t|t}}^{-1} &= \widehat{\mathbf{\Lambda}}_{t|t-1}^{-1} + \mathbf{C}^T \mathbf{R}^{-1} \mathbf{C}
\end{aligned} \quad (23)$$

The update equation of PCF as presented in [19] are as follows:

$$\begin{aligned}
\hat{\mathbf{x}}_{\mathbf{n}_{t|t}} &= \hat{\mathbf{x}}_{t|t-1} + \widehat{\mathbf{\Lambda}}_{\mathbf{n}_{t|t}} \widetilde{\mathbf{C}}^T \widetilde{\mathbf{R}}^{-1} [\mathbf{n}_t - \hat{\mathbf{n}}_{t|t-1}] \\
\widehat{\mathbf{\Lambda}}_{\mathbf{n}_{t|t}}^{-1} &= \widehat{\mathbf{\Lambda}}_{t|t-1}^{-1} + \widetilde{\mathbf{C}}^T \widetilde{\mathbf{R}}^{-1} \widetilde{\mathbf{C}} \\
\widetilde{\mathbf{C}} &= \left(\widehat{\mathbf{\Lambda}}_{t|t-1}^{-1} \mathbf{\Lambda}_{\mathbf{xn}}\right)^T \\
\widetilde{\mathbf{R}} &= \mathbf{\Lambda}_{\mathbf{nn}} - \widetilde{\mathbf{C}} \widehat{\mathbf{\Lambda}}_{t|t-1} \widetilde{\mathbf{C}}^T
\end{aligned} \quad (24)$$

The necessary terms are computed using the 5$^{th}$-degree spherical radial cubature rule [49] as follows:

$$\begin{aligned}
\hat{\mathbf{n}}_{t|t-1} &\approx \sum_{a=1}^{2d^2+1} w_a E[\mathbf{n}_t|\mathbf{x}_{t,a}] \\
\mathbf{\Lambda_{xn}} &\approx \sum_{a=1}^{2d^2+1} w_a \left(\mathbf{x}_{t,a} E[\mathbf{n}_t|\mathbf{x}_{t,a}]^T\right) - \hat{\mathbf{x}}_{t|t-1}\hat{\mathbf{n}}_{t|t-1}^T \\
\mathbf{\Lambda_{nn}} &\approx \sum_{a=1}^{2d^2+1} w_a \left(V[\mathbf{n}_t|\mathbf{x}_{t,a}] + E[\mathbf{n}_t|\mathbf{x}_{t,a}]E[\mathbf{n}_t|\mathbf{x}_{t,a}]^T\right) - \hat{\mathbf{n}}_{t|t-1}\hat{\mathbf{n}}_{t|t-1}^T \\
E[\mathbf{n}_t|\mathbf{x}_{t,a}] &= [\lambda_1(\mathbf{x}_{t,a})\Delta, \cdots, \lambda_c(\mathbf{x}_{t,a})\Delta] \\
V[\mathbf{n}_t|\mathbf{x}_{t,a}] &= \text{diag}(\lambda_1(\mathbf{x}_{t,a})\Delta, \cdots, \lambda_c(\mathbf{x}_{t,a})\Delta) \\
\mathbf{x}_{t,a} &= \hat{\mathbf{x}}_{t|t-1} + \sqrt{\hat{\mathbf{\Lambda}}_{t|t-1}}\xi_a \\
w_1 &= \frac{2}{d+2} \\
w_{2:2d+1} &= \frac{4-d}{2(d+2)^2} \\
w_{2d+2:2d^2+1} &= \frac{1}{(d+2)^2} \\
\xi_1 &= 0 \\
\xi_{2:2d+1} &= \{\pm\sqrt{d+2} * e_j : j = 1, 2, \ldots, d\} \\
\xi_{2d+2:2d^2+1} &= \{\pm\sqrt{d+2} * \frac{1}{\sqrt{2}}(e_j \pm e_k) : j < k, j, k = 1, 2, \ldots, d\}
\end{aligned} \quad (25)$$

## 7.2. MSNF information fusion

Here we present the detailed steps in deriving (10) from (9). First is the covariance equation. Substituting (23) and (24) into (9), we have:

$$\begin{aligned}
\hat{\mathbf{\Lambda}}_{t|t}^{-1} &= \hat{\mathbf{\Lambda}}_{\mathbf{n}_{t|t}}^{-1} + \hat{\mathbf{\Lambda}}_{\mathbf{y}_{t|t}}^{-1} - \hat{\mathbf{\Lambda}}_{t|t-1}^{-1} \\
&= \hat{\mathbf{\Lambda}}_{t|t-1}^{-1} + \tilde{\mathbf{C}}^T\tilde{\mathbf{R}}^{-1}\tilde{\mathbf{C}} + \hat{\mathbf{\Lambda}}_{t|t-1}^{-1} + \mathbf{C}^T\mathbf{R}^{-1}\mathbf{C} - \hat{\mathbf{\Lambda}}_{t|t-1}^{-1} \\
&= \hat{\mathbf{\Lambda}}_{t|t-1}^{-1} + \tilde{\mathbf{C}}^T\tilde{\mathbf{R}}^{-1}\tilde{\mathbf{C}} + \mathbf{C}^T\mathbf{R}^{-1}\mathbf{C}
\end{aligned} \quad (26)$$

Then we derive the mean equation in (9) by substituting (23) and (24) into it again.

$$\begin{aligned}
\hat{\mathbf{x}}_{\mathbf{t}|\mathbf{t}} &= \hat{\mathbf{\Lambda}}_{t|t}\left(\hat{\mathbf{\Lambda}}_{\mathbf{n}_{t|t}}^{-1}\hat{\mathbf{x}}_{\mathbf{n}_{t|t}} + \hat{\mathbf{\Lambda}}_{\mathbf{y}_{t|t}}^{-1}\hat{\mathbf{x}}_{\mathbf{y}_{t|t}} - \hat{\mathbf{\Lambda}}_{t|t-1}^{-1}\hat{\mathbf{x}}_{\mathbf{t}|\mathbf{t-1}}\right) \\
&= \hat{\mathbf{\Lambda}}_{t|t}\left(\hat{\mathbf{\Lambda}}_{\mathbf{n}_{t|t}}^{-1}\left(\hat{\mathbf{x}}_{t|t-1} + \hat{\mathbf{\Lambda}}_{\mathbf{n}_{t|t}}\tilde{\mathbf{C}}^T\tilde{\mathbf{R}}^{-1}[\mathbf{n}_t - \hat{\mathbf{n}}_{t|t-1}]\right) + \hat{\mathbf{\Lambda}}_{\mathbf{y}_{t|t}}^{-1}\left(\hat{\mathbf{x}}_{t|t-1} + \hat{\mathbf{\Lambda}}_{\mathbf{y}_{t|t}}\mathbf{C}^T\mathbf{R}^{-1}[\mathbf{y}_t - \mathbf{C}\hat{\mathbf{x}}_{t|t-1}]\right) - \hat{\mathbf{\Lambda}}_{t|t-1}^{-1}\hat{\mathbf{x}}_{t|t-1}\right) \\
&= \hat{\mathbf{\Lambda}}_{t|t}\left[\left(\hat{\mathbf{\Lambda}}_{\mathbf{n}_{t|t}}^{-1} + \hat{\mathbf{\Lambda}}_{\mathbf{y}_{t|t}}^{-1} - \hat{\mathbf{\Lambda}}_{t|t-1}^{-1}\right)\hat{\mathbf{x}}_{t|t-1} + \tilde{\mathbf{C}}^T\tilde{\mathbf{R}}^{-1}[\mathbf{n}_t - \hat{\mathbf{n}}_{t|t-1}] + \mathbf{C}^T\mathbf{R}^{-1}[\mathbf{y}_t - \mathbf{C}\hat{\mathbf{x}}_{t|t-1}]\right] \\
&= \hat{\mathbf{x}}_{t|t-1} + \hat{\mathbf{\Lambda}}_{t|t}\tilde{\mathbf{C}}^T\tilde{\mathbf{R}}^{-1}[\mathbf{n}_t - \hat{\mathbf{n}}_{t|t-1}] + \hat{\mathbf{\Lambda}}_{t|t}\mathbf{C}^T\mathbf{R}^{-1}[\mathbf{y}_t - \mathbf{C}\hat{\mathbf{x}}_{t|t-1}]
\end{aligned} \quad (27)$$

Note that the last line in (27) comes from $\hat{\mathbf{\Lambda}}_{t|t}^{-1} = \hat{\mathbf{\Lambda}}_{\mathbf{n}_{t|t}}^{-1} + \hat{\mathbf{\Lambda}}_{\mathbf{y}_{t|t}}^{-1} - \hat{\mathbf{\Lambda}}_{t|t-1}^{-1}$ in (9).

## 7.3. Complete equations for sMSNF

First, we show equations for the three steps in (12) to compute $f\left(\mathbf{x}_t \middle| \mathbf{h}_{1:t}, s_t^{(j)}\right)$ from $f\left(\mathbf{x}_{t-1} \middle| \mathbf{h}_{1:t-1}, s_{t-1}^{(j)}\right)$.

**Step 1**: $f\left(\mathbf{x}_{t-1}|\mathbf{h}_{1:t-1}, s_{t-1}^{(i)}\right) \to f\left(\mathbf{x}_{t-1}|\mathbf{h}_{1:t-1}, s_t^{(j)}\right)$. We approximate $f\left(\mathbf{x}_{t-1}|\mathbf{h}_{1:t-1}, s_t^{(j)}\right)$ as Gaussian and denote its mean and covariance are $\hat{\mathbf{x}}_{t-1|t-1}^{(-,j)}$ and $\widehat{\boldsymbol{\Lambda}}_{t-1|t-1}^{(-,j)}$, respectively. Then we can compute it by [20]

$$
\begin{aligned}
P\left(s_{t-1}^{(i)}\big|s_t^{(j)}, \mathbf{h}_{1:t-1}\right) &\propto P\left(s_t^{(j)}\big|s_{t-1}^{(i)}\right) P(s_{t-1}^{(i)}|\mathbf{h}_{1:t-1}) \\
\hat{\mathbf{x}}_{t-1|t-1}^{(-,j)} &= \sum_{i=1}^M \hat{\mathbf{x}}_{t-1|t-1}^{(i)} P(s_{t-1}^{(i)}|s_t^{(j)}, \mathbf{h}_{1:t-1}) \\
\widehat{\boldsymbol{\Lambda}}_{t-1|t-1}^{(-,j)} &= \sum_{i=1}^M \left[\widehat{\boldsymbol{\Lambda}}_{t-1|t-1}^{(i)} + \left(\hat{\mathbf{x}}_{t-1|t-1}^{(i)} - \hat{\mathbf{x}}_{t-1|t-1}^{(-,j)}\right)\left(\hat{\mathbf{x}}_{t-1|t-1}^{(i)} - \hat{\mathbf{x}}_{t-1|t-1}^{(-,j)}\right)^T\right] P(s_{t-1}^{(i)}|s_t^{(j)}, \mathbf{h}_{1:t-1})
\end{aligned}
\tag{28}
$$

**Step 2**: $f\left(\mathbf{x}_{t-1}|\mathbf{h}_{1:t-1}, s_t^{(j)}\right) \to f\left(\mathbf{x}_t|\mathbf{h}_{1:t-1}, s_t^{(j)}\right)$. This is the prediction (6) in MSNF conditioned on $s_t^{(j)}$.

$$
\begin{aligned}
\hat{\mathbf{x}}_{t|t-1}^{(j)} &= \mathbf{A}^{(j)} \hat{\mathbf{x}}_{t-1|t-1}^{(-,j)} \\
\widehat{\boldsymbol{\Lambda}}_{t|t-1}^{(j)} &= \mathbf{A}^{(j)} \widehat{\boldsymbol{\Lambda}}_{t-1|t-1}^{(-,j)} \mathbf{A}^{(j)T} + \mathbf{Q}^{(j)}
\end{aligned}
\tag{29}
$$

**Step 3**: $f\left(\mathbf{x}_t|\mathbf{h}_{1:t-1}, s_t^{(j)}\right) \to f\left(\mathbf{x}_t|\mathbf{h}_{1:t}, s_t^{(j)}\right)$. This is the update step (10) in MSNF conditioned on $s_t^{(j)}$.

$$
\begin{aligned}
\left(\widehat{\boldsymbol{\Lambda}}_{t|t}^{(j)}\right)^{-1} &= \left(\widehat{\boldsymbol{\Lambda}}_{t|t-1}^{(j)}\right)^{-1} + \tilde{\mathbf{C}}^{(j)T}\widetilde{\mathbf{R}}^{(j)-1}\tilde{\mathbf{C}}^{(j)} + \mathbf{C}^{(j)T}\mathbf{R}^{(j)-1}\mathbf{C}^{(j)} \\
\hat{\mathbf{x}}_{t|t}^{(j)} &= \hat{\mathbf{x}}_{t|t-1}^{(j)} + \widehat{\boldsymbol{\Lambda}}_{t|t}^{(j)}\tilde{\mathbf{C}}^{(j)T}\widetilde{\mathbf{R}}^{(j)-1}\left[\mathbf{n}_t - \hat{\mathbf{n}}_{t|t-1}^{(j)}\right] + \widehat{\boldsymbol{\Lambda}}_{t|t}^{(j)}\mathbf{C}^{(j)T}\mathbf{R}^{(j)-1}\left[\mathbf{y}_t - \mathbf{C}^{(j)}\hat{\mathbf{x}}_{t|t-1}^{(j)}\right] \\
\tilde{\mathbf{C}}^{(j)} &= \left(\widehat{\boldsymbol{\Lambda}}_{t|t-1}^{(j)-1}\boldsymbol{\Lambda}_{\mathbf{xn}}^{(j)}\right)^T \\
\widetilde{\mathbf{R}}^{(j)} &= \boldsymbol{\Lambda}_{\mathbf{nn}}^{(j)} - \tilde{\mathbf{C}}^{(j)}\widehat{\boldsymbol{\Lambda}}_{t|t-1}^{(j)}\tilde{\mathbf{C}}^{(j)T}
\end{aligned}
\tag{30}
$$

Second, we compute $P\left(s_t^{(j)}\big|\mathbf{h}_{1:t}\right)$ from $P(s_{t-1}^{(i)}|\mathbf{h}_{1:t-1})$ following the sMSF in [20] as follows:

$$
\begin{aligned}
P\left(s_t^{(j)}\big|\mathbf{h}_{1:t-1}\right) &= \sum_{i=1}^M P\left(s_t^{(j)}\big|s_{t-1}^{(i)}\right) P(s_{t-1}^{(i)}|\mathbf{h}_{1:t-1}) \\
P\left(s_t^{(j)}\big|\mathbf{h}_{1:t}\right) &\propto P\left(s_t^{(j)}\big|\mathbf{h}_{1:t-1}\right) f\left(\mathbf{h}_t\big|\mathbf{h}_{1:t-1}, s_t^{(j)}\right) \\
f\left(\mathbf{h}_t\big|\mathbf{h}_{1:t-1}, s_t^{(j)}\right) &= f\left(\mathbf{n}_t\big|\hat{\mathbf{x}}_{t|t}^{(j)}, s_t^{(j)}\right) f\left(\mathbf{y}_t\big|\hat{\mathbf{x}}_{t|t}^{(j)}, s_t^{(j)}\right)^\tau \sqrt{\det\widehat{\boldsymbol{\Lambda}}_{t|t}^{(j)} / \det\widehat{\boldsymbol{\Lambda}}_{t|t-1}^{(j)}} \\
&\quad \times \exp\left(-\frac{1}{2}\left(\hat{\mathbf{x}}_{t|t}^{(j)} - \hat{\mathbf{x}}_{t|t-1}^{(j)}\right)^T \widehat{\boldsymbol{\Lambda}}_{t|t-1}^{(j)-1}\left(\hat{\mathbf{x}}_{t|t}^{(j)} - \hat{\mathbf{x}}_{t|t-1}^{(j)}\right)\right)
\end{aligned}
\tag{31}
$$

After obtaining terms for each regime, latent state estimates of each regime are combined as follows:

$$
\begin{aligned}
\hat{\mathbf{x}}_{t|t} &= \sum_{j=1}^M \hat{\mathbf{x}}_{t|t}^{(j)} P(s_t^{(j)}|\mathbf{h}_{1:t}) \\
\widehat{\boldsymbol{\Lambda}}_{t|t} &= \sum_{j=1}^M \left[\widehat{\boldsymbol{\Lambda}}_{t|t}^{(j)} + \left(\hat{\mathbf{x}}_{t|t}^{(j)} - \hat{\mathbf{x}}_{t|t}\right)\left(\hat{\mathbf{x}}_{t|t}^{(j)} - \hat{\mathbf{x}}_{t|t}\right)^T\right] P(s_t^{(j)}|\mathbf{h}_{1:t})
\end{aligned}
\tag{32}
$$

*7.4. Switching multiscale smoother*

Here, we present the equations for switching multiscale smoother (SMS). The goal of SMS is to estimate the mean and the covariance of $f(\mathbf{x}_{t-1}|\mathbf{h}_{1:T})$, denoted by $\hat{\mathbf{x}}_{t-1|T}$, and $\widehat{\mathbf{\Lambda}}_{t-1|T}$ from the estimates provided by switching multiscale filters, such as sMSF or sMSNF. The final smoother equations are as follows:

$$\hat{\mathbf{x}}_{t-1|T}^{(i,j)} = \hat{\mathbf{x}}_{t-1|t-1}^{(i)} + \mathbf{J}_{t-1}^{(i,j)}\left(\hat{\mathbf{x}}_{t|T}^{(j)} - \hat{\mathbf{x}}_{t|t-1}^{(j)}\right)$$

$$\widehat{\mathbf{\Lambda}}_{t-1|T}^{(i,j)} = \widehat{\mathbf{\Lambda}}_{t-1|t-1}^{(i)} + \mathbf{J}_{t-1}^{(i,j)}\left(\widehat{\mathbf{\Lambda}}_{t|T}^{(j)} - \widehat{\mathbf{\Lambda}}_{t|t-1}^{(j)}\right)\mathbf{J}_{t-1}^{(i,j)T}$$

$$\mathbf{J}_{t-1}^{(i,j)} = \widehat{\mathbf{\Lambda}}_{t-1|t-1}^{(i)}\mathbf{A}^{(j)T}\widehat{\mathbf{\Lambda}}_{t|t-1}^{(j)^{-1}}$$

$$P\left(s_{t-1}^{(i)}, s_t^{(j)}\middle|\mathbf{h}_{1:T}\right) = P(s_{t-1}^{(i)}|s_t^{(j)}, \mathbf{h}_{1:T})P(s_t^{(j)}|\mathbf{h}_{1:T}) \approx P(s_{t-1}^{(i)}|s_t^{(j)}, \mathbf{h}_{1:t-1})P(s_t^{(j)}|\mathbf{h}_{1:T})$$

$$P(s_{t-1}^{(i)}|\mathbf{h}_{1:T}) = \sum_{j=1}^{M} P\left(s_{t-1}^{(i)}, s_t^{(j)}\middle|\mathbf{h}_{1:T}\right)$$

$$P\left(s_t^{(j)}\middle|s_{t-1}^{(i)}, \mathbf{h}_{1:T}\right) = \frac{P(s_{t-1}^{(i)}, s_t^{(j)}|\mathbf{h}_{1:T})}{P(s_{t-1}^{(i)}|\mathbf{h}_{1:T})}$$

$$\hat{\mathbf{x}}_{t-1|T}^{(i)} = \sum_{j=1}^{M} \hat{\mathbf{x}}_{t-1|T}^{(i,j)} P(s_t^{(j)}|s_{t-1}^{(i)}, \mathbf{h}_{1:T})$$

$$\widehat{\mathbf{\Lambda}}_{t-1|T}^{(i)} = \sum_{j=1}^{M}\left[\widehat{\mathbf{\Lambda}}_{t-1|T}^{(i,j)} + \left(\hat{\mathbf{x}}_{t-1|T}^{(i,j)} - \hat{\mathbf{x}}_{t-1|T}^{(i)}\right)\left(\hat{\mathbf{x}}_{t-1|T}^{(i,j)} - \hat{\mathbf{x}}_{t-1|T}^{(i)}\right)^T\right] P(s_t^{(j)}|s_{t-1}^{(i)}, \mathbf{h}_{1:T})$$

$$\hat{\mathbf{x}}_{t-1|T} = \sum_{j=1}^{M} \hat{\mathbf{x}}_{t-1|T}^{(i)} P(s_{t-1}^{(i)}|\mathbf{h}_{1:T})$$

$$\widehat{\mathbf{\Lambda}}_{t-1|T} = \sum_{j=1}^{M}\left[\widehat{\mathbf{\Lambda}}_{t-1|T}^{(i)} + \left(\hat{\mathbf{x}}_{t-1|T}^{(i)} - \hat{\mathbf{x}}_{t-1|T}\right)\left(\hat{\mathbf{x}}_{t-1|T}^{(i)} - \hat{\mathbf{x}}_{t-1|T}\right)^T\right] P(s_{t-1}^{(i)}|\mathbf{h}_{1:T})$$

(33)

The details of the derivation are presented in [20].

### 7.5. Likelihood scaling parameter propagation

Here we detail how the switching multiscale filters are affected by the likelihood scaling parameter. From (13), we know that $f(\mathbf{h}_t|\mathbf{x}_t, s_t) = f(\mathbf{n}_t|\mathbf{x}_t, s_t) \times f(\mathbf{y}_t|\mathbf{x}_t, s_t)^\tau$, so we redo (7) again with condition $s_t^{(j)}$ as follow:

$$f\left(\mathbf{x}_t \mid \mathbf{h}_{1:t}, s_t^{(j)}\right) = \frac{f\left(\mathbf{h}_t \mid \mathbf{x}_t, \mathbf{h}_{1:t-1}, s_t^{(j)}\right) f\left(\mathbf{x}_t \mid \mathbf{h}_{1:t-1}, s_t^{(j)}\right)}{f\left(\mathbf{h}_t \mid \mathbf{h}_{1:t-1}, s_t^{(j)}\right)}$$

$$= f\left(\mathbf{n}_t \mid \mathbf{x}_t, \mathbf{h}_{1:t-1}, s_t^{(j)}\right) \times f\left(\mathbf{y}_t \mid \mathbf{x}_t, \mathbf{h}_{1:t-1}, s_t^{(j)}\right)^\tau \times \frac{f\left(\mathbf{x}_t \mid \mathbf{h}_{1:t-1}, s_t^{(j)}\right)}{f\left(\mathbf{h}_t \mid \mathbf{h}_{1:t-1}, s_t^{(j)}\right)}$$

$$= \frac{f\left(\mathbf{x}_t \mid \mathbf{n}_t, \mathbf{h}_{1:t-1}, s_t^{(j)}\right) f\left(\mathbf{n}_t \mid \mathbf{h}_{1:t-1}, s_t^{(j)}\right)}{f\left(\mathbf{x}_t \mid \mathbf{h}_{1:t-1}, s_t^{(j)}\right)} \times \frac{f\left(\mathbf{x}_t \mid \mathbf{y}_t, \mathbf{h}_{1:t-1}, s_t^{(j)}\right)^\tau f\left(\mathbf{y}_t \mid \mathbf{h}_{1:t-1}, s_t^{(j)}\right)^\tau}{f\left(\mathbf{x}_t \mid \mathbf{h}_{1:t-1}, s_t^{(j)}\right)^\tau}$$

$$\times \frac{f\left(\mathbf{x}_t \mid \mathbf{h}_{1:t-1}, s_t^{(j)}\right)}{f\left(\mathbf{h}_t \mid \mathbf{h}_{1:t-1}, s_t^{(j)}\right)}$$

$$\propto \frac{f\left(\mathbf{x}_t \mid \mathbf{n}_t, \mathbf{h}_{1:t-1}, s_t^{(j)}\right) \times f\left(\mathbf{x}_t \mid \mathbf{y}_t, \mathbf{h}_{1:t-1}, s_t^{(j)}\right)^\tau}{f\left(\mathbf{x}_t \mid \mathbf{h}_{1:t-1}, s_t^{(j)}\right)^\tau} \quad (34)$$

$$\propto \exp\left(-\frac{1}{2}\left(\left(\mathbf{x}_t - \hat{\mathbf{x}}_{\mathbf{n}_t|t}^{(j)}\right)^T \hat{\mathbf{\Lambda}}_{\mathbf{n}_t|t}^{(j)^{-1}} \left(\mathbf{x}_t - \hat{\mathbf{x}}_{\mathbf{n}_t|t}^{(j)}\right) + \tau \left(\mathbf{x}_t - \hat{\mathbf{x}}_{\mathbf{y}_t|t}^{(j)}\right)^T \hat{\mathbf{\Lambda}}_{\mathbf{y}_t|t}^{(j)^{-1}} \left(\mathbf{x}_t - \hat{\mathbf{x}}_{\mathbf{y}_t|t}^{(j)}\right)\right.\right.$$

$$\left.\left. - \tau \left(\mathbf{x}_t - \hat{\mathbf{x}}_{t|t-1}^{(j)}\right)^T \hat{\mathbf{\Lambda}}_{t|t-1}^{(j)^{-1}} \left(\mathbf{x}_t - \hat{\mathbf{x}}_{t|t-1}^{(j)}\right)\right)\right)$$

$$\propto \exp\left(-\frac{1}{2} \mathbf{x}_t^T \left(\hat{\mathbf{\Lambda}}_{\mathbf{n}_t|t}^{(j)^{-1}} + \tau \hat{\mathbf{\Lambda}}_{\mathbf{y}_t|t}^{(j)^{-1}} - \tau \hat{\mathbf{\Lambda}}_{t|t-1}^{(j)^{-1}}\right) \mathbf{x}_t\right.$$

$$\left. + \mathbf{x}_t^T \left(\hat{\mathbf{\Lambda}}_{\mathbf{n}_t|t}^{(j)^{-1}} \hat{\mathbf{x}}_{\mathbf{n}_t|t}^{(j)} + \tau \hat{\mathbf{\Lambda}}_{\mathbf{y}_t|t}^{(j)^{-1}} \hat{\mathbf{x}}_{\mathbf{y}_t|t}^{(j)} - \tau \hat{\mathbf{\Lambda}}_{t|t-1}^{(j)^{-1}} \hat{\mathbf{x}}_{t|t-1}^{(j)}\right)\right)$$

So, the update mean, and covariance are expressed as:

$$\left(\hat{\mathbf{\Lambda}}_{t|t}^{(j)}\right)^{-1} = \left(\hat{\mathbf{\Lambda}}_{\mathbf{n}_t|t}^{(j)}\right)^{-1} + \tau \left(\hat{\mathbf{\Lambda}}_{\mathbf{y}_t|t}^{(j)}\right)^{-1} - \tau \left(\hat{\mathbf{\Lambda}}_{t|t-1}^{(j)}\right)^{-1}$$

$$\hat{\mathbf{x}}_{t|t}^{(j)} = \hat{\mathbf{\Lambda}}_{t|t} \left(\left(\hat{\mathbf{\Lambda}}_{\mathbf{n}_t|t}^{(j)}\right)^{-1} \hat{\mathbf{x}}_{\mathbf{n}_t|t}^{(j)} + \tau \left(\hat{\mathbf{\Lambda}}_{\mathbf{y}_t|t}^{(j)}\right)^{-1} \hat{\mathbf{x}}_{\mathbf{y}_t|t}^{(j)} - \tau \left(\hat{\mathbf{\Lambda}}_{t|t-1}^{(j)}\right)^{-1} \hat{\mathbf{x}}_{t|t-1}^{(j)}\right) \quad (35)$$

By following the steps of (26) and (27), we yield the update equation presented in (14) from (35).

There is an additional term that is required for estimating the regime state probability that is affected by the likelihood scaling parameter. As outlined in [20], one of steps in (31) of calculating the regime distribution $P\left(s_t^{(j)} \mid \mathbf{h}_{1:t}\right)$ is to estimate $f\left(\mathbf{h}_t \mid \mathbf{h}_{1:t-1}, s_t^{(j)}\right)$. We obtain this term by using the prediction and update densities as follows:

$$f\left(\mathbf{h}_t \mid \mathbf{h}_{1:t-1}, s_t^{(j)}\right) = \frac{f\left(\mathbf{h}_t \mid \mathbf{x}_t, \mathbf{h}_{1:t-1}, s_t^{(j)}\right) f\left(\mathbf{x}_t \mid \mathbf{h}_{1:t-1}, s_t^{(j)}\right)}{f\left(\mathbf{x}_t \mid \mathbf{h}_{1:t}, s_t^{(j)}\right)}$$

$$= f\left(\mathbf{n}_t \mid \mathbf{x}_t, \mathbf{h}_{1:t-1}, s_t^{(j)}\right) f\left(\mathbf{y}_t \mid \mathbf{x}_t, \mathbf{h}_{1:t-1}, s_t^{(j)}\right)^\tau \frac{f\left(\mathbf{x}_t \mid \mathbf{h}_{1:t-1}, s_t^{(j)}\right)}{f\left(\mathbf{x}_t \mid \mathbf{h}_{1:t}, s_t^{(j)}\right)} \quad (36)$$

$$= f\left(\mathbf{n}_t \mid \hat{\mathbf{x}}_{t|t}^{(j)}, s_t^{(j)}\right) f\left(\mathbf{y}_t \mid \hat{\mathbf{x}}_{t|t}^{(j)}, s_t^{(j)}\right)^\tau \sqrt{\det \widehat{\mathbf{\Lambda}}_{t|t}^{(j)} / \det \widehat{\mathbf{\Lambda}}_{t|t-1}^{(j)}}$$

$$\times \exp\left(-\frac{1}{2}\left(\hat{\mathbf{x}}_{t|t}^{(j)} - \hat{\mathbf{x}}_{t|t-1}^{(j)}\right)^T \widehat{\mathbf{\Lambda}}_{t|t-1}^{(j)}{}^{-1} \left(\hat{\mathbf{x}}_{t|t}^{(j)} - \hat{\mathbf{x}}_{t|t-1}^{(j)}\right)\right)$$

### 7.6. E-Step equations

Here, we present the additional equations needed for the E-step for switching multiscale dynamical systems. In the expected complete log-likelihood (18), all the probability terms are computed using the switching multiscale smoother in appendix 7.4. The remaining terms are computed as follows:

$$E\left[\cdot \mid \mathbf{h}_{1:T}, s_t^{(j)}\right] \triangleq \langle \cdot \rangle^{(j)}$$

$$\langle \mathbf{x}_t \mathbf{x}_t^T \rangle^{(j)} = \hat{\mathbf{x}}_{t|T}^{(j)} \left(\hat{\mathbf{x}}_{t|T}^{(j)}\right)^T + \widehat{\mathbf{\Lambda}}_{t|T}$$

$$\mathbf{J}_{t-1}^{(i,j)} = \widehat{\mathbf{\Lambda}}_{t-1|t-1}^{(i)} \mathbf{A}^{(j)T} \widehat{\mathbf{\Lambda}}_{t|t-1}^{(j)}{}^{-1}$$

$$\langle \mathbf{x}_t \mathbf{x}_{t-1}^T \rangle^{(j)} = \hat{\mathbf{x}}_{t|T}^{(j)} \left(\hat{\mathbf{x}}_{t-1|T}^{(j)}\right)^T + \widehat{\mathbf{\Lambda}}_{t|T} \mathbf{J}_{t-1}^{(i,j)T}$$

$$E[\log N(\mathbf{x}_0; \boldsymbol{\mu}_0, \mathbf{\Lambda}_0) \mid \mathbf{h}_{1:T}] = -\frac{1}{2} tr\left(\mathbf{\Lambda}_0^{-1}(\widehat{\mathbf{\Lambda}}_{0|T} - \hat{\mathbf{x}}_{0|T}\boldsymbol{\mu}_0^T - \boldsymbol{\mu}_0\hat{\mathbf{x}}_{0|T}^T + \boldsymbol{\mu}_0\boldsymbol{\mu}_0^T)\right) - \frac{1}{2}\log|\mathbf{\Lambda}_0| + const$$

$$E\left[\log N\left(\mathbf{x}_t; \mathbf{A}^{(j)}\mathbf{x}_{t-1}, \mathbf{Q}^{(j)}\right) \mid \mathbf{h}_{1:T}, s_t^{(j)}\right]$$

$$= -\frac{1}{2} tr\left((\mathbf{Q}^{(j)})^{-1} \sum_{t=1}^T \left(\langle \mathbf{x}_t \mathbf{x}_t^T \rangle^{(j)} - \mathbf{A}^{(j)}\langle \mathbf{x}_{t-1}\mathbf{x}_t^T \rangle^{(j)} - \langle \mathbf{x}_t\mathbf{x}_{t-1}^T \rangle^{(j)}(\mathbf{A}^{(j)})^T \right.\right. \quad (37)$$

$$\left.\left. + \mathbf{A}^{(j)}\langle \mathbf{x}_{t-1}\mathbf{x}_{t-1}^T \rangle^{(j)}(\mathbf{A}^{(j)})^T\right)\right) - \frac{T}{2}\log|\mathbf{Q}^{(j)}| + const$$

$$E\left[\log Poisson\left(n_t^c; \exp\left(\alpha_c^{(j)} + \boldsymbol{\beta}_c^{(j)T}\mathbf{x}_t\right)\right) \mid \mathbf{h}_{1:T}, s_t^{(j)}\right]$$

$$= \sum_{t=1}^T \left(n_t^c(\alpha_c + \boldsymbol{\beta}_c^T\hat{\mathbf{x}}_{t|T}^{(j)}) - \exp\left(\alpha_c + \boldsymbol{\beta}_c^T\hat{\mathbf{x}}_{t|T}^{(j)} + \frac{1}{2}\boldsymbol{\beta}_c^T\widehat{\mathbf{\Lambda}}_{t|T}^{(j)}\boldsymbol{\beta}_c\right)\right)$$

$$E\left[\log N\left(\mathbf{y}_t; \mathbf{C}^{(j)}\mathbf{x}_t, \mathbf{R}^{(j)}\right) \mid \mathbf{h}_{1:T}, s_t^{(j)}\right]$$

$$= -\frac{1}{2} tr\left((\mathbf{R}^{(j)})^{-1} \sum_{t=1}^T \left(\mathbf{y}_t\mathbf{y}_t^T - \mathbf{C}^{(j)}\hat{\mathbf{x}}_{t|T}^{(j)}\mathbf{y}_t^T - \mathbf{y}_t\hat{\mathbf{x}}_{t|T}^{(j)}(\mathbf{C}^{(j)})^T + \mathbf{C}^{(j)}\langle \mathbf{x}_t\mathbf{x}_t^T \rangle^{(j)}(\mathbf{C}^{(j)})^T\right)\right)$$

$$- \frac{T}{2}\log|\mathbf{R}^{(j)}| + const$$

where $tr(\cdot)$ is the matrix trace.

### 7.7. M-Step equations

Here, we present the equations for the M-step for switching multiscale dynamical systems. Since the parameters of the regime, the latent state, the continuous and discrete neural signals are separated in the expected complete log-likelihood (18), the following M-step equations are the same as in [19] and [20] except $C^{(j)^*}$ and $R^{(j)^*}$, which are also derived in the same manner. Note that stationary dynamical system equations can also be computed using the equations by setting the number of regimes $M = 1$. We denote the new set of parameters with superscript $*$.

$$
\begin{aligned}
E[\cdot | \mathbf{h}_{1:T}] &\triangleq \langle \cdot \rangle \\
E\left[\cdot \middle| \mathbf{h}_{1:T}, s_t^{(j)}\right] &\triangleq \langle \cdot \rangle^{(j)} \\
W_t^{(j)} &= P\left(s_t^{(j)} \middle| \mathbf{h}_{1:T}\right) \\
\mathbf{A}^{(j)^*} &= \left(\sum_{t=1}^T W_t^{(j)} \langle \mathbf{x}_t \mathbf{x}_{t-1}^T \rangle^{(j)}\right) \left(\sum_{t=1}^T W_t^{(j)} \langle \mathbf{x}_{t-1} \mathbf{x}_{t-1}^T \rangle^{(j)}\right)^{-1} \\
\mathbf{Q}^{(j)^*} &= \left(\sum_{t=1}^T W_t^{(j)} \left(\langle \mathbf{x}_t \mathbf{x}_t^T \rangle^{(j)} - \mathbf{A}^{(j)^*} \langle \mathbf{x}_{t-1} \mathbf{x}_t^T \rangle^{(j)} - \langle \mathbf{x}_t \mathbf{x}_{t-1}^T \rangle^{(j)} \mathbf{A}^{(j)^* T} \right.\right. \\
&\quad \left.\left. + \mathbf{A}^{(j)^*} \langle \mathbf{x}_{t-1} \mathbf{x}_{t-1}^T \rangle^{(j)} \mathbf{A}^{(j)^* T}\right)\right) \left(\sum_{t=1}^T W_t^{(j)}\right)^{-1} \\
\boldsymbol{\mu}_0^* &= \langle \mathbf{x}_0 \rangle \\
\boldsymbol{\Lambda}_0^* &= \langle \mathbf{x}_0 \mathbf{x}_0^T \rangle - \langle \mathbf{x}_0 \rangle \langle \mathbf{x}_0 \rangle^T \\
\boldsymbol{\Phi}_{j,i}^* &= \left(\sum_{t=2}^T P\left(s_t^{(j)}, s_{t-1}^{(i)} \middle| \mathbf{h}_{1:T}\right)\right) \left(\sum_{t=1}^{T-1} W_t^{(j)}\right)^{-1} \\
\pi_j^* &= W_1^{(j)} \\
\mathbf{C}^{(j)^*} &= \left(\sum_{t=1}^T W_t^{(j)} \mathbf{y}_t \langle \mathbf{x}_t^T \rangle^{(j)}\right) \left(\sum_{t=1}^T W_t^{(j)} \langle \mathbf{x}_t \mathbf{x}_t^T \rangle^{(j)}\right)^{-1} \\
\mathbf{R}^{(j)^*} &= \left(\sum_{t=1}^T W_t^{(j)} \left(\mathbf{y}_t \mathbf{y}_t^T - \mathbf{C}^{(j)^*} \langle \mathbf{x}_t \rangle^{(j)} \mathbf{y}_t^T - \mathbf{y}_t \langle \mathbf{x}_t^T \rangle^{(j)} \mathbf{C}^{(j)^* T} \right.\right. \\
&\quad \left.\left. + \mathbf{C}^{(j)^*} \langle \mathbf{x}_t \mathbf{x}_t^T \rangle^{(j)} \mathbf{C}^{(j)^* T}\right)\right) \left(\sum_{t=1}^T W_t^{(j)}\right)^{-1}
\end{aligned}
\tag{38}
$$

For the Poisson observation equation parameters, we use the trust-region methods [50] in MATLAB.

$$
\begin{aligned}
\begin{bmatrix} \alpha_c^{(j)^*} \\ \boldsymbol{\beta}_c^{(j)^*} \end{bmatrix} &= \underset{[\alpha_c; \boldsymbol{\beta}_c]}{\operatorname{argmax}} \sum_{t=1}^T W_t^{(j)} \langle n_t^c (\alpha_c + \boldsymbol{\beta}_c^T \mathbf{x}_t) - \exp(\alpha_c + \boldsymbol{\beta}_c^T \mathbf{x}_t) \rangle^{(j)} \\
&= \underset{[\alpha_c; \boldsymbol{\beta}_c]}{\operatorname{argmax}} \sum_{t=1}^T W_t^{(j)} \left( n_t^c (\alpha_c + \boldsymbol{\beta}_c^T \hat{\mathbf{x}}_{t|T}^{(j)}) \right. \\
&\quad \left. - \exp\left(\alpha_c + \boldsymbol{\beta}_c^T \hat{\mathbf{x}}_{t|T}^{(j)} + \frac{1}{2} \boldsymbol{\beta}_c^T \hat{\boldsymbol{\Lambda}}_{t|T}^{(j)} \boldsymbol{\beta}_c\right) \right)
\end{aligned}
\tag{39}
$$